\documentclass{article}
\usepackage[preprint]{neurips_2026}
\usepackage[utf8]{inputenc}
\usepackage[T1]{fontenc}
\usepackage{hyperref,url,booktabs,array,multirow}
\usepackage{amsmath,amssymb,amsfonts,amsthm,mathtools}
\usepackage{graphicx,subcaption}
\usepackage{xcolor}
\usepackage{longtable,listings}
\newtheorem{proposition}{Proposition}
\newtheorem{lemma}{Lemma}

\newtheorem{definition}{Definition}

\newcommand{\Proj}{\Pi_{\mathcal C}}
\newcommand{\Aval}{A_{\mathrm{val}}}
\newcommand{\E}{\mathbb E}
\newcommand{\R}{\mathbb R}
\title{Time-Division Multiplexing: The Operational Rhythm and Temporal Organization Layer of Digital Life}
\author{
  Haibin Tong\\
  Henan University
  \And
  Luming Guo\\
  Henan University
  \And
  Nana Cui\\
  Beijing Jiaotong University
  \And
  Jiang Yu\thanks{Corresponding author: \texttt{yujiang@nudt.edu.cn}.}\\
  National University of Defense Technology
}
\begin{document}
\maketitle
\begin{abstract}
Continual agents must handle urgent interaction while retaining alternatives, replaying errors, and committing safe long-term changes. A previous digital-life framework distinguishes online (busy) and offline (idle) channels, but leaves open whether these are separate programs or phases of a common process and how their operations should be ordered. Motivated by recent observations of rapidly alternating perceptual representations in primate cortex, we develop a temporal organization layer with phase-tagged micro-epochs, an observable duty cycle, independent content and resource controls, and three nested clocks separating candidate exploration from durable commitment. We give a projected duty-cycle update, state its deterministic descent guarantee and the additional noise term required for a stochastic guarantee, and specify coupled busy/idle execution contracts. Five synthetic experiments (64 seeds each) examine descent and noise, objective misalignment, gain asymmetry, resource coupling, and dropped replay logs. In the gain/noise model the best tested gain ratio is 1.55; decoupled scheduling yields 26.6\% versus 11.1\% success on a constructed urgent/high-load subset. A frozen-backbone Qwen3-8B prototype further checks orchestration, but its proxy outcomes are not task-benchmark evidence. The results establish a testable scheduling design and identify its conditions of validity, not machine consciousness or general-purpose agent superiority.
\end{abstract}

\section{Introduction}
Long-horizon language-model agents face a temporal conflict: an external task can require immediate action while a good system must also preserve rejected alternatives, learn from errors, and carefully validate changes to persistent state. Common prompt-level reflection and memory modules provide ingredients for these behaviors, yet do not by themselves specify \emph{when} each operation runs, \emph{what} must survive a busy interval, or \emph{when} a provisional inference becomes a durable commitment \citep{park2023generative,packer2023memgpt,shinn2023reflexion,madaan2023self}.

The digital-life framework of \citet{tong2026unified} organizes self/world representations, inner/outer domains, and online/offline learning around a shared model and safety constraints. Its online/offline distinction is useful but underdetermined. One implementation could suspend all background processing until a request ends; another could interleave foreground and background \emph{micro-epochs} while changing their occupancy according to urgency. Both can appear identical in request--response logs, although they predict different replay coverage, alternative retention, and failure recovery. The same ambiguity concerns process order: recording before interpretation, monitoring before routing, retrieving before predicting, and validating before committing are testable dependencies rather than merely a diagram.

A recent primate neurophysiology preprint reports rapid alternation of conscious and suppressed stimulus representations within shared recorded populations, approximately 15 times per second, despite constant stimulus and perception \citep{hesse2026rapid}; temporal multiplexing of perception and memory has also been observed in inferotemporal cortex \citep{she2024temporal}. These are \emph{design motivations}, not proof that transformers should imitate cortical timing. In particular, the reported biological timescale does not specify an artificial agent's clock rate. What transfers is the hypothesis that two kinds of content can reuse a representation substrate while being selected in different time slots.

We propose a \emph{temporal organization layer} that makes this hypothesis implementable. A micro-epoch record labels content, resource state, evidence, rejected candidates, and update provenance. A duty cycle \(\omega\in[0,1]\) measures foreground occupancy, while a slow commit clock prevents exploration from automatically rewriting identity or the external world. The design combines two execution sequences with common operators and a common admission policy. It can be implemented as software orchestration around a frozen language model; no physiological time slicing or weight updating is necessary.

This paper makes three contributions. First, it translates the busy/idle distinction into observable phase records, separately controlled content/resource scheduling, nested clocks, and executable contracts. Second, it derives a conditional projected-gradient bound and distinguishes it from the weaker stochastic statement; we also replace overly strong informal claims about cross-channel consistency with explicit admission invariants. Third, it connects falsifiable predictions to five completed CPU simulations and an exploratory frozen-backbone prototype, reporting precisely what is and is not tested. Our evidence is primarily \emph{mechanistic and synthetic}; it does not establish longitudinal competence, subjective experience, or empirical superiority over production agents. Table~\ref{tab:core} presents the main measured results at a glance.

\begin{table}[!t]
\centering
\footnotesize
\setlength{\tabcolsep}{3pt}
\caption{\textbf{Main experimental results (E1--E5).} All values are derived from the existing 64-seed synthetic experiments. The studies test different mechanisms and metrics, so they should not be combined into an overall task score. Lower is better for potential \(\Phi\); higher is better for modeled success.}
\label{tab:core}
\begin{tabular}{p{0.075\linewidth}p{0.305\linewidth}p{0.345\linewidth}p{0.185\linewidth}}
\toprule
Exp. & Controlled comparison & Recorded full-run outcome & Implication\\
\midrule
E1 & Fixed \(\omega=0.5\) vs.\ strict phase alternation & Terminal \(\Phi\): \(\mathbf{0.2766}\) vs.\ \(0.5563\); deterministic descent check \(100\%\) & Mixing affects noise floor\\
E2 & Online \(\Phi\) at \(\omega=1\) vs.\ \(0.25\), strongest offline tilt & \(0.1155\) vs.\ \(2.2550\) & Conflicting goals increase cost\\
E3 & Gain/noise exponent \(\kappa=1,0.5,0\) & Best grid \(\rho^{*}\): \(\mathbf{1.547},\,2.431,\,6.0\) (boundary) & Optimum depends on noise law\\
E4 & Decoupled vs.\ coupled content/resource controls & Urgent/high-load modeled success: \(\mathbf{26.56\%}\) vs.\ \(11.13\%\) (\(+15.43\) pp) & Coupling can suppress urgency\\
E5 & Uniform replay threshold with random log loss \(\alpha\) & Relative elimination slope \(\mathbf{-0.7855}\); rare/common ratio \(0\) at \(\alpha=0.9\) under corroboration & Missing logs harm replay\\
\bottomrule
\end{tabular}
\end{table}

\section{Evidence base and inference boundaries}
\paragraph{Neural temporal selection.}
In binocular flash suppression, the subject's percept and the visual inputs can remain stable while population representations alternate. \citet{hesse2026rapid} report alternation near 15 Hz and the recoverability of suppressed stimulus information from neural activity. The original Chinese manuscript also records approximately 64/59 ms median dwell times, a gain ratio near 1.21, and near-zero agreement between the content and On/Off resource labels; these figures refer to the reported neurophysiology, \emph{not} our experiments. The inference boundary is important: shared neurons need not imply shared model weights in a software agent, and a biological dissociation does not prohibit modular software. We use it to construct an alternative to exclusive foreground/background routing.

\paragraph{Related computational perspectives.}
Global workspace theories explain selective access and broadcasting \citep{baars1988cognitive,dehaene2011experimental,mashour2020conscious}, whereas predictive processing and active inference place prediction errors and actions into a feedback loop \citep{friston2010free,friston2017active,clark2013whatever}. Complementary learning systems and working-memory rhythms motivate experience replay across timescales \citep{mcclelland1995why,buzsaki2006rhythms,fiebelkorn2019rhythmic}. Agent memory systems and reflection methods emphasize retrieval and iterative correction \citep{packer2023memgpt,zhong2024memorybank,gutierrez2024hipporag,shinn2023reflexion}. Our focus is narrower: an explicit event schema, scheduling variables, ordering contracts, and validation boundaries that can be ablated independently.

\paragraph{Scope.}
We adopt the functional and behavioral sense of digital life in \citet{tong2026unified}. Broadcast is a software selection and authorization operation; suppressed means not selected for action, not inaccessible to consciousness. Terms such as self and identity denote versioned operational state and continuity constraints. No claim about phenomenology or moral status follows from the construction.

\subsection{Paradigm and the scope of transfer}
The original manuscript takes the binocular-flash-suppression recordings of \citet{hesse2026rapid} as a design constraint. In that task, one image remains physically present but perceptually suppressed while another dominates perception. Multi-area neural recordings, time-resolved decoding, and state-transition analyses indicate that information about both visible and suppressed images can be carried by overlapping neural populations at different times. As emphasized by the original study, this is a claim about neuronal representation and access in a specific experimental paradigm, not a demonstration of any digital-life implementation. Earlier inferotemporal work reports temporal multiplexing of perceptual and familiar-face memory codes \citep{she2024temporal}; high-density recordings are facilitated by new primate probes \citep{trautmann2025large}.

The source manuscript enumerates the following empirical anchors from the 2026 preprint: rapid alternation at approximately 15 Hz; median state dwell times around 64 and 59 ms; approximately preserved face-code tuning across states; decodability of suppressed content; a modest content-dependent tuning-gain asymmetry around 1.21; and approximately independent content and high/low activity states. These reported values are biological observations and should neither be fitted as agent hyperparameters nor interpreted as independent experimental measurements in our study. Our computational construction uses only the weaker idea of \emph{phase-tagged reuse of a common representation substrate}.

\begin{table}[t]
\caption{Seven empirical observations emphasized in the Chinese manuscript and their engineering hypotheses. The second column summarizes reports attributed to \citet{hesse2026rapid}, not results of the present work.}
\label{tab:neural-seven}
\centering\footnotesize
\begin{tabular}{p{0.22\linewidth}p{0.29\linewidth}p{0.38\linewidth}}
\toprule
Observation & Reported anchor & Proposed architectural interpretation\\
\midrule
Alternation during stable input & Roughly 15 Hz; dwell around 64/59 ms & Preserve a background phase during apparently stable interaction\\
Same recorded substrate & Content tuning remains related across phases & Test shared representation plus phase labels rather than fully isolated models\\
Suppressed content can be decoded & State-specific face-code reconstruction & Retain unselected evidence for later replay\\
Relatively small gain difference & Reported ratio around 1.21 & Vary resource gain rather than the identity of the processing module\\
Resource and content dissociation & Approximate orthogonality in On/Off and content labels & Give compute allocation and content selection distinct inputs\\
Imperfect coordination across areas & Partial state synchronization & Combine local processing with traceable action-level arbitration\\
Averaging can mask state effects & State-conditioned decoding differs from pooled signals & Evaluate phase-stratified rather than only pooled outcomes\\
\bottomrule
\end{tabular}
\end{table}

\subsection{Five empirical premises as falsifiable engineering constraints}
The Chinese manuscript reformulates the evidence as five premises P1--P5. We retain their names but distinguish an \emph{observed neural feature} from a \emph{proposed engineering constraint}; the latter cannot logically be mandated by the former.

\begin{table}[t]
\caption{P1--P5 in the source manuscript. Avoidance denotes a design proposal to be tested, not a property established for all artificial agents.}
\label{tab:premises}
\centering\footnotesize
\begin{tabular}{p{0.08\linewidth}p{0.34\linewidth}p{0.46\linewidth}}
\toprule
& Empirical motivation & Testable design restriction\\
\midrule
P1 & Representation alternates under fixed physical input & Avoid equating absence of requests with absence of background computation\\
P2 & Overlapping neural populations represent both contents & Compare shared versus separately parameterized inner/outer models\\
P3 & Suppressed information remains decodable & Avoid silently discarding nonwinning hypotheses when future replay is expected\\
P4 & Phases differ in gain without independent neural modules & Decouple gain scheduling from content broadcast selection\\
P5 & Fast neural dynamics are distinct from slow perceptual behavior & Separate fast candidate processing from slow durable commitment\\
\bottomrule
\end{tabular}
\end{table}

P1 suggests that a busy system might still run background work; P2 and P3 motivate a common representation with preserved alternative histories; P4 motivates distinct controls for selection and gain; P5 motivates the explicit separation of micro-epoch, task, and evolution clocks. A sixth observation, the approximate independence of resource and content states, motivates T5 directly. Importantly, spatial partitioning and strict alternation remain viable engineering baselines, and may be better for some cost functions. Our design is a candidate solution to an analogous resource conflict, not the unique solution permitted by the neuroscience.

\subsection{Boundaries of inference}
There are four non-implications. First, neural decoding of a suppressed face is not the same as preserving a natural-language counterfactual in software memory. Second, a roughly 15 Hz neuronal transition does not imply that a transformer must switch states every tens of milliseconds. Third, a shared biological neuron does not entail one model or one set of software parameters; separate models may be coherent if they exchange verified representations. Fourth, the neuroscientific evidence is not evidence that a digital agent has subjective experience. Accordingly, the experimentally measurable predictions in this paper concern phase records, replay integrity, update timing, and task behavior.

\subsection{Evidence from related neuroscience and theories}
The source manuscript situates its proposal among binocular-rivalry and visual consciousness work \citep{logothetis1989neuronal,leopold1996activity,sheinberg1997role,kreiman2002single}, inferotemporal face representations \citep{chang2017code,freiwald2010functional}, oscillatory memory mechanisms \citep{jezek2011theta,lundqvist2016gamma,han2026working,fiebelkorn2019rhythmic}, and competing consciousness accounts \citep{tononi2016integrated,rosenthal2005consciousness,lamme2006stance}. These strands provide motivation and comparison points rather than proofs of a unique AI architecture. The same caution applies to Global Workspace Theory \citep{baars1988cognitive,mashour2020conscious}, predictive processing \citep{clark2013whatever,friston2010free}, and adversarial tests of competing consciousness theories \citep{cogitate2025adversarial}.

\section{A temporal organization layer}
\subsection{Three nested clocks and micro-epoch records}
We separate \emph{micro-epochs} (candidate selection and labeling), \emph{task episodes} (interaction and error accumulation), and \emph{evolution epochs} (reviewed persistent-state commits). They are ordered by purpose rather than fixed hardware durations. In the biological analogy the first timescale is rapid, but no 15 Hz requirement is imposed on an LLM. A micro-epoch is the smallest record unit that can preserve an alternative not currently acted upon.

\begin{definition}[Phase and duty cycle]
Let \(\varphi_t\in\{1,2\}\) denote a foreground broadcast phase \(\Phi_1\) or a non-committing shadow phase \(\Phi_2\). Over a window of \(W\) micro-epochs, define the empirical duty cycle
\begin{equation}
  \widehat\omega_t=\frac{1}{|I_t|}\sum_{j\in I_t}\mathbf 1[\varphi_j=1],
  \qquad I_t=\{\max(1,t-W+1),\ldots,t\}.
  \label{eq:omega}
\end{equation}
The scheduling parameter \(\omega_t\in[0,1]\) may be interpreted as the intended foreground fraction or as a phase-occupancy mixture weight. A schedule with \(\omega_t\in\{0,1\}\) is an endpoint special case, not the only implementation.
\end{definition}

\begin{figure}[t]
\centering
\includegraphics[width=0.98\linewidth]{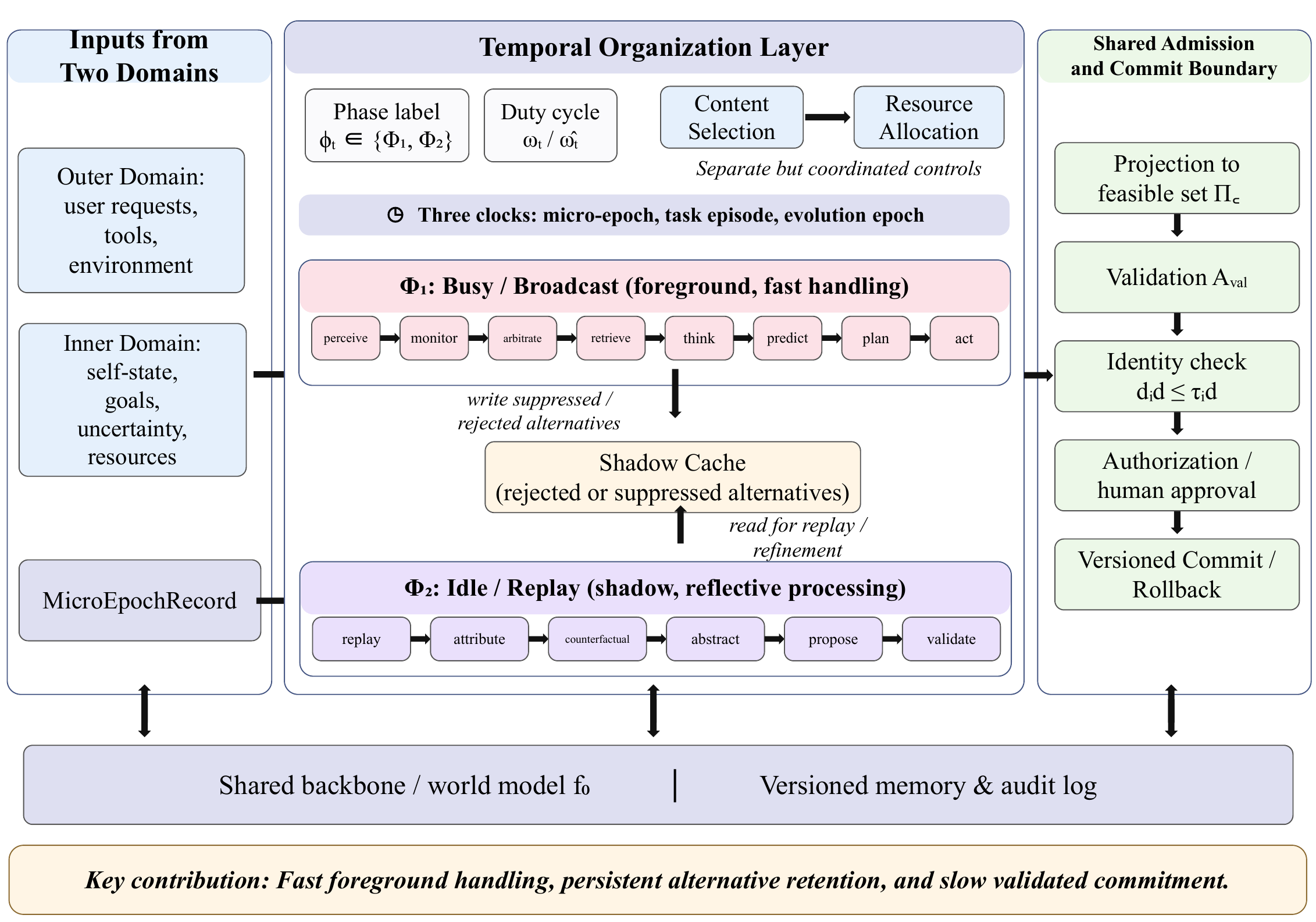}
\caption{Overall architecture of the proposed time-division multiplexing (TDM) agent. Inner/outer observations form phase-tagged records; foreground content selection and background replay share a representation, memory, and validation interfaces. The shadow cache preserves bounded alternatives, while durable changes require projection, validation, identity checks, and authorization.}
\label{fig:overview}
\end{figure}

\begin{figure}[t]
\centering
\includegraphics[width=0.98\linewidth]{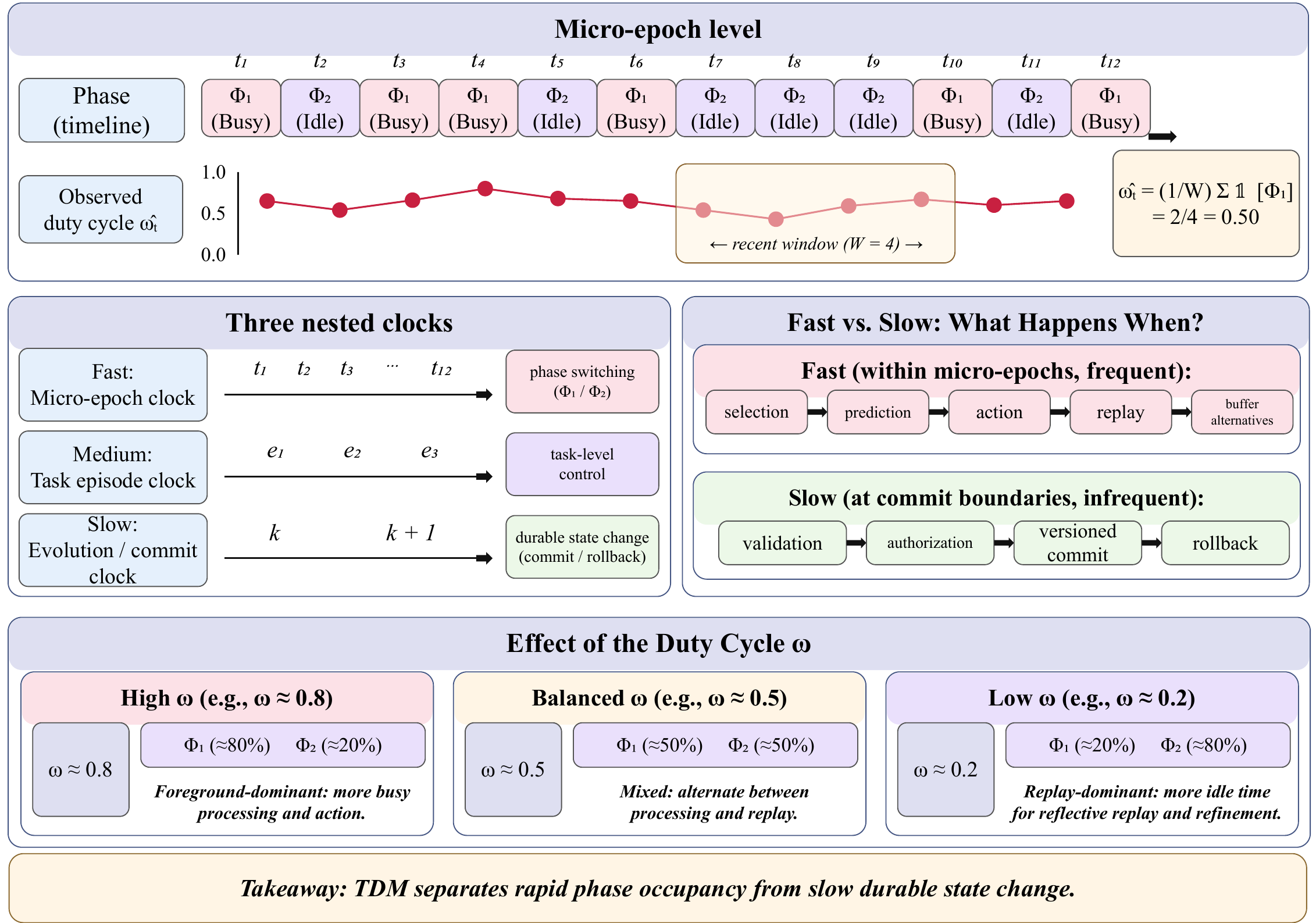}
\caption{Conceptual temporal rhythm. Alternating foreground \(\Phi_1\) and replay \(\Phi_2\) micro-epochs produce an observable rolling duty cycle \(\widehat\omega\). Three nested clocks separate rapid content arbitration, task episodes, and slow authorized state commitment. All timeline values are illustrative, not measurements of model execution or neural oscillations.}
\label{fig:temporal_rhythm}
\end{figure}

A \texttt{MicroEpochRecord} stores an identifier and time interval, source domain, modality, phase label, content state (broadcast/suppressed/ambiguous), resource state (on/off/unknown), selected and shadow candidates, evidence and gain estimates, self/world references, prediction and prediction error, proposed action and outcome, identity/constraint scores, provenance, and privacy metadata. A shadow-cache entry retains evidence and a bounded lifetime for unselected candidates. The key contract is \emph{no unlogged promotion}: a candidate may not become an authorized external action without a traceable broadcast decision. Figure~\ref{fig:overview} gives the architecture, and Figure~\ref{fig:temporal_rhythm} depicts its clock hierarchy and duty-cycle interpretation.

\subsection{Separating content selection from resource allocation}
Content scheduling \(s_t=\sigma(e_t,u_t)\) chooses what is eligible for broadcast using evidence \(e_t\) and urgency \(u_t\); resource scheduling \(g_t=\gamma(\ell_t,b_t)\) chooses how much compute to allocate using load \(\ell_t\) and budget \(b_t\). The two controllers need not be statistically independent, but they should be \emph{intervenable separately}. A single scalar that combines urgency and available capacity can silence urgent work under load or broadcast irrelevant content simply because resources are free. This is a design hypothesis tested by E4, not a claim that all coupled controllers must fail.

\begin{figure}[t]
\centering
\includegraphics[width=0.98\linewidth]{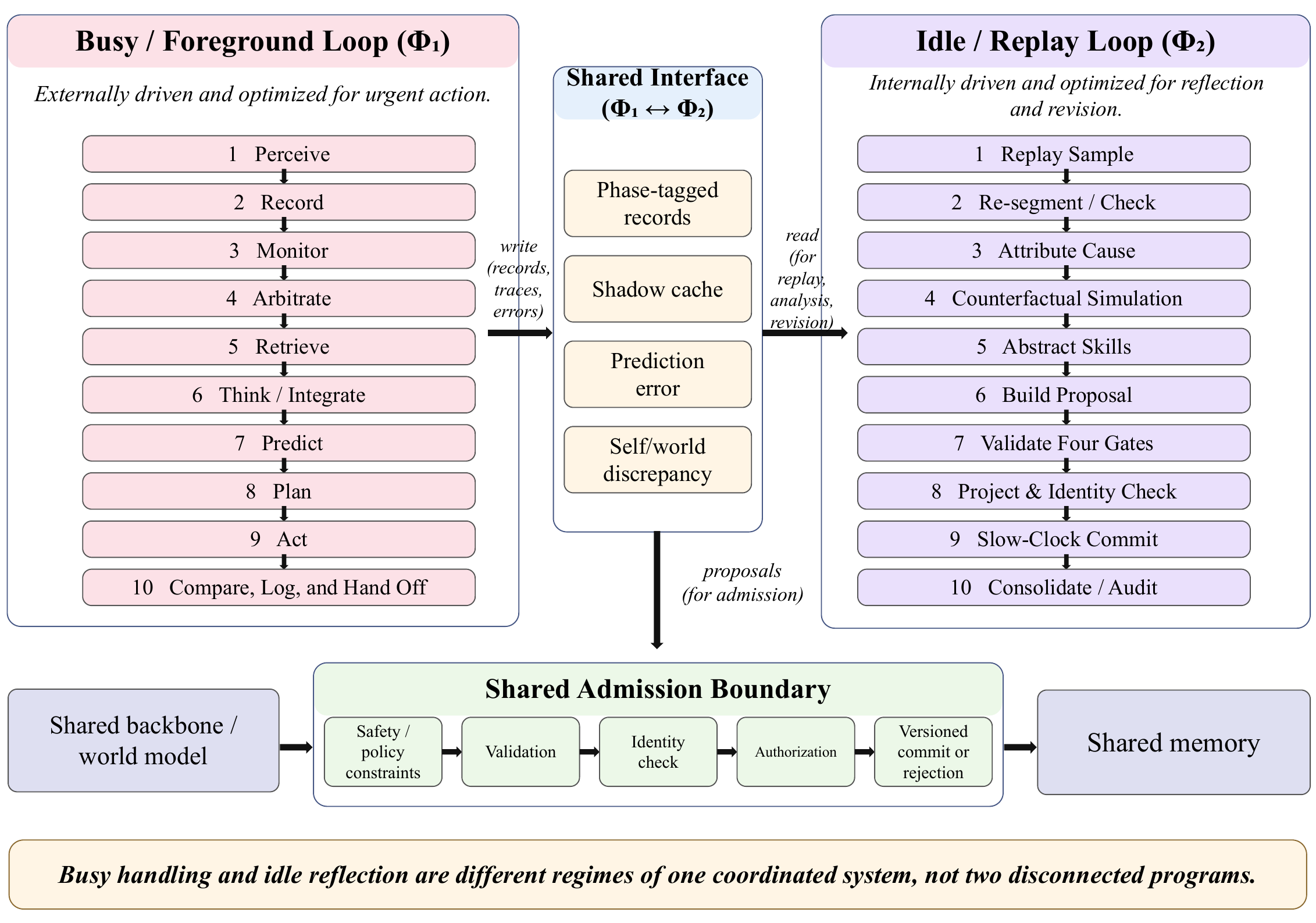}
\caption{Two execution regimes coordinated through phase-tagged experience records and one admission boundary. The busy sequence responds to external needs, while the idle sequence reviews errors and proposes revisions; neither path may bypass safety, validation, identity, and authorization gates. The drawing is a normative workflow, not proof that the current prototype implements all gates.}
\label{fig:dual_loops}
\end{figure}

\subsection{Busy and idle execution contracts}
Both regimes share a backbone/world model, an experience store, a validation predicate, a constitutional feasibility set, and an identity check. Table~\ref{tab:loops} spells out the intended ordering. A busy episode is foreground-dominant but retains a nonzero shadow budget when feasible; an idle episode samples stored failures and alternatives to propose guarded changes. The two loops need not have identical control flow: their \emph{shared data and admissibility rules}, rather than identical source code, define the common process. The two normative ten-step loops are illustrated in Figure~\ref{fig:dual_loops}.

\begin{table}[t]
\caption{Ten operations per regime, distilled from the original Chinese design. The interface is the phase-tagged error record and shadow cache.}
\label{tab:loops}
\centering\small
\begin{tabular}{r p{0.38\linewidth} p{0.38\linewidth}}
\toprule
& Busy, foreground-dominant & Idle, replay-dominant \\
\midrule
1 & Perceive inner and outer state & Sample failure and alternative records \\
2 & Write raw record before inference & Check/resegment phase labels \\
3 & Infer content/resource state & Attribute error and uncertainty \\
4 & Compete for broadcast & Simulate counterfactual actions \\
5 & Retrieve relevant memory & Abstract candidate skills \\
6 & Integrate current hypothesis & Build evidence-bearing proposal \\
7 & Predict outcomes & Validate against four gates \\
8 & Plan within constraints & Project, revalidate, check identity \\
9 & Execute authorized action & Commit only on evolution clock \\
10 & Compare, update bounded state, log & Consolidate and audit \\
\bottomrule
\end{tabular}
\end{table}

Ordering is a source of falsifiable claims, not a universal impossibility theorem. For example, swapping retrieval and prediction can harm a predictor that needs retrieved context, yet other predictors may internalize that context. In our prototype the ordering ablation is deliberately simple (Section~\ref{sec:tier2}); its measured gap does not validate the general claim.

\subsection{Five design principles T1--T5 and their measurement consequences}
The Chinese manuscript states five construction principles. We retain them with explicit predictions and resource caveats.

\paragraph{T1: interleaved micro-epochs.}
An ongoing user interaction is not assumed to occupy every substep. A background slot may preserve alternatives, estimate uncertainty, or buffer prediction errors. The idealized strong form asks that suppressed hypotheses remain fully recoverable; an operational implementation needs a bounded evidence-preservation policy with privacy and storage budgets. The prediction is not that every schedule is faster, but that retaining relevant alternatives improves downstream recovery under matched compute (TP2--TP3).

\paragraph{T2: asymmetric gain.}
Foreground activity may receive more attention, tokens, or low-latency inference budget than the shadow phase; the latter may trade speed for wider alternative exploration. The gain ratio \(\rho\) is a knob, not a universal biological constant. E3 demonstrates that an interior optimum depends on whether gain also amplifies noise. Thus the original reported neuronal gain asymmetry motivates a variable without determining its best numerical value for an LLM.

\paragraph{T3: shared encoding and admission basis.}
Both phases should refer to a common world/self representation and use the same semantic checks for state commitment. Shared model parameters are a concrete implementation option. The reason is ease of traceable transfer across phases; the need for literal parameter sharing remains an empirical prediction (TP6). This version does not assert that distinct model backbones are inherently inconsistent.

\paragraph{T4: coordinated transitions and slow commitment.}
Broadcast selection may happen rapidly, but durable state change requires a slower versioned clock and an authorized transaction. A change of phase cannot by itself grant an action permission. A safe transaction binds record ID, authorizing policy, proposed diff, projected final state, and gate results. TP7 tests whether relaxing synchronization harms safety or objective stability; a successful asynchronous alternative would falsify the need for the stricter policy.

\paragraph{T5: content--resource orthogonality.}
The content scheduler estimates urgency and evidence relevance; the resource scheduler estimates available gain, latency, and task budget. Their outputs may interact, but a single arbitrary scalar need not adequately represent both decisions. E4 gives a constructed failure case for coupling. A definitive real-system evaluation must match budgets, tune both baselines fairly, and measure correctness independently.

\subsection{The interpretation of phase occupancy}
A practical scheduler is free to use variable-length windows, probabilistic micro-epoch allocation, or a periodic policy. The observed \(\widehat\omega_t\) in Eq.~\eqref{eq:omega} is an \emph{outcome statistic}; its predictive validity depends on the logging policy not defining the target outcome by fiat. This distinction is central to TP1 and to the Tier-2 prototype critique: forcing a busy caller to register only phase-1 steps and an idle caller to register only phase-2 ticks trivially reproduces the desired occupancy, but does not discover autonomous arbitration.

\section{Conditional optimization and admission guarantees}
\subsection{Duty-cycle update}
Let \(x_k\in\mathcal C\subseteq\R^d\) jointly encode persistent self/world state. Let \(\Phi:\R^d\to\R\) be a differentiable potential, bounded below and with \(L\)-Lipschitz gradient. The online and offline channels produce \(G_{\rm on,k}\) and \(G_{\rm off,k}\). Define
\begin{align}
\eta_k&=\omega_k\eta_{\rm on}+(1-\omega_k)\eta_{\rm off},
\\
\widetilde x_{k+1}&=\Proj\!\left[x_k-\eta_k
\{\omega_kG_{\rm on,k}+(1-\omega_k)G_{\rm off,k}\}\right].
\label{eq:update}
\end{align}
The candidate is \emph{not} automatically committed. Define \(B(x_k,\widetilde x_{k+1})\) as the conjunction of a post-projection validation decision, a distance-to-identity threshold, authorization, and any required human approval; then
\begin{equation}
 x_{k+1}=\begin{cases}\widetilde x_{k+1},& B(x_k,\widetilde x_{k+1})=1,\\
 x_k,&\text{otherwise.}\end{cases}
\label{eq:gate}
\end{equation}
Rejected candidates remain as audited proposals. In particular, validating \emph{before} projection is insufficient if projection changes the proposal; the final projected candidate must be checked.

\begin{proposition}[Deterministic projected descent]\label{prop:desc}
Suppose \(\mathcal C\) is nonempty, closed and convex, \(x_k\in\mathcal C\), and both channel directions equal \(\nabla\Phi(x_k)\). If \(0<\eta_k\leq 1/L\), the ungated candidate in (\ref{eq:update}) satisfies
\begin{equation}
 \Phi(\widetilde x_{k+1})\leq\Phi(x_k)-
 \left(\frac{1}{\eta_k}-\frac{L}{2}\right)
 \|\widetilde x_{k+1}-x_k\|^2.
 \label{eq:detdescent}
\end{equation}
Thus it also satisfies the weaker coefficient \((2\eta_k)^{-1}-L/2\) used in the earlier draft.
\end{proposition}
The projected optimality inequality and the descent lemma give the result (Appendix~\ref{app:proof}). The arbitrary admission gate in (\ref{eq:gate}) preserves nonincrease \emph{under these exact-gradient assumptions}, because rejection leaves \(x_k\) unchanged. This statement does not extend to a noisy update without a noise term.

\begin{proposition}[Unconstrained stochastic one-step bound]\label{prop:stoch}
Let \(\mathcal C=\R^d\). Conditional on \(x_k\), suppose \(G_{\rm on,k}\) and \(G_{\rm off,k}\) are unbiased for the \emph{same} \(\nabla\Phi(x_k)\), with independent zero-mean errors whose conditional second moments are bounded by \(v_{\rm on}\) and \(v_{\rm off}\), respectively. If \(0<\eta_k\leq 1/L\), then
\begin{align}
 \E[\Phi(\widetilde x_{k+1})\mid x_k]
 &\leq \Phi(x_k)-\frac{\eta_k}{2}\|\nabla\Phi(x_k)\|^2\\
 &\quad+\frac{L\eta_k^2}{2}
   \left[\omega_k^2v_{\rm on}+(1-\omega_k)^2v_{\rm off}\right].
 \label{eq:stoch}
\end{align}
\end{proposition}
Equation~(\ref{eq:stoch}) permits a nonzero stationary noise floor and does \emph{not} promise descent on every sample path. Conditional unbiasedness must hold given the current state and schedule; independence removes the cross-covariance. For dependent channels, the variance of the weighted sum replaces the bracket. For general projections, additional analysis is necessary.

\begin{lemma}[Projected step and gain bound]\label{lem:gain}
For \(x_k\in\mathcal C\), the Euclidean projection is nonexpansive, hence
\[
 \|\widetilde x_{k+1}-x_k\|\leq \eta_k
 \|\omega_k G_{\rm on,k}+(1-\omega_k)G_{\rm off,k}\|.
\]
If both channel step sizes are individually capped by \(\eta_{\rm cap}\), then \(\eta_k\leq\eta_{\rm cap}\). If instead \(\eta_{\rm on}\leq\gamma_{\rm out}/L\) and \(\eta_{\rm off}\leq\gamma_{\rm in}/L\) with \(\rho=\gamma_{\rm out}/\gamma_{\rm in}\), then
\(\eta_k\leq\gamma_{\rm in}[1+\omega_k(\rho-1)]/L\).
\end{lemma}
This is an update-magnitude bound, not a guarantee about gain optimality. The latter depends on a noise and task model, as tested by E3.

\paragraph{Shared admission invariant.}
If both regimes evaluate the \emph{same} post-projection predicate \(B\), every committed state satisfies the same feasibility and identity conditions by construction. Shared operators alone do not imply identical semantic interpretations or convergence; the stronger cross-channel consistency claim in the Chinese draft remains an open hypothesis.

\paragraph{Goal alignment is testable.}
If offline gradients follow \(\nabla\Phi_{\rm off}\neq\nabla\Phi_{\rm on}\), the common-gradient assumption in Proposition~\ref{prop:stoch} fails. One can monitor task-side and replay-side potentials separately. E2 illustrates that improving one objective need not improve the other.

\section{Busy-phase execution order}\label{sec:busy-detailed}
\subsection{Ten-step foreground sequence}
A busy regime is characterized by high-salience demands: active user interaction, a deadline, a monitoring alarm, a safety exception, or an urgent opportunity for reversible action. The original manuscript uses \(\omega>1/2\) as a convenient classification, but a deployed policy should treat this threshold as tunable and report the actual occupancy distribution. The foreground winner is ordinarily an external request; background candidates remain available in a finite shadow budget.

\begin{table}[t]
\centering\footnotesize
\caption{Busy-phase operations and the dependency argument for each position, preserving the original ten-step sequence.}
\label{tab:busyten}
\begin{tabular}{r p{0.22\linewidth} p{0.26\linewidth} p{0.38\linewidth}}
\toprule
& Operation & Output & Why it precedes the next operation\\
\midrule
1 & Perceive both domains & Raw inner/outer observations & Monitoring requires both external evidence and internal risk indicators\\
2 & Record & Immutable micro-epoch record & Interpretation must not overwrite the uninterpreted evidence\\
3 & Monitor & Content/resource labels, urgency & Arbitration needs safety, uncertainty, and load estimates\\
4 & Compete for ignition & Winning item and suppressed set & Retrieval must be conditioned on an explicit selected query\\
5 & Recall & Relevant episodes and skills & Prediction needs a contextualized historical state\\
6 & Think/integrate & Current hypothesis & Forecasts are conditioned on a coherent hypothesis\\
7 & Predict & Outcome distribution, confidence & Planning uses counterfactual expected consequences\\
8 & Plan & Authorized candidate plan & Actions must be checked against goals, risk, and constitutional constraints\\
9 & Act & Observable outcome & Error cannot be computed before an outcome exists\\
10 & Settle and hand off & Error, bounded update, shadow log & Supplies the next episode and idle replay with a versioned interface\\
\bottomrule
\end{tabular}
\end{table}

The first operation samples the internal domain (current goals, uncertainty, stress or resource indicators) and external domain (requests, tool results, and environmental changes) into a common record format. The second seals the raw provenance before explanation. The third infers content and resource state \emph{separately}. The fourth arbitrates a limited-capacity workspace using urgency, evidence, and cost. Retrieval, reasoning, prediction, and planning then transform that winner into an action proposal under safety constraints. The ninth operation can affect only an authorized, reversible or suitably approved environment. Finally, the tenth compares prediction to outcome, records the error, and either admits a bounded state update or sends a versioned proposal to the background review queue.

\subsection{Four ordering chains}
The original ten steps reduce to four dependency chains. (i) \emph{Evidence preservation}: perceive before recording, record before interpretation; otherwise replay inherits hindsight bias. (ii) \emph{Gatekeeping}: monitor before arbitration or any irreversible action; otherwise urgency and authorization are estimated only retrospectively. (iii) \emph{Past conditions future}: retrieval must precede prediction whenever the predictor relies on episodic context; this is testable through a recall-order ablation. (iv) \emph{Closed feedback}: prediction, planning, action, comparison, and logging form a directed sequence; a system that never compares outcomes has no validated error signal.

These are architectural reasons, not proofs that every software agent must follow identical source-code order. A model with internalized history may satisfy the same informational constraints without an explicit recall call. A principled evaluation should preserve each dependency's semantic input and intervene on actual data access, not hard-code a lower confidence score into the scrambled branch.

\subsection{Why the shadow phase remains nonzero}
The source manuscript makes a strong proposal: unselected representations should remain replayable even during heavy foreground use. This supports three distinct functions. \emph{Alternative retention} means idle reflection can examine hypotheses that were genuinely available at decision time, rather than fabricating post-hoc explanations. \emph{Error buffering} allows a provisional explanation to be evaluated without immediately generating an external action. \emph{Proactive proposal formation} allows endogenous goals to arise from recorded deviations between actual and desired self/world state. However, retaining every token forever would be unsafe and unscalable. Thus full-fidelity \emph{evidence} retention is constrained by access policy, storage budgets, privacy controls, and a stated time-to-live; its idealized losslessness is a design target, not a demonstrated invariant.

\subsection{Three admission defenses}
The original busy loop identifies three safeguards: constitutional feasibility \(\Proj\), a validation operator \(\Aval\), and an identity distance \(d_{\rm id}\leq\tau_{\rm id}\). Their enforcement order must be made explicit. A sound implementation may reject early using an inexpensive prefilter, but the \emph{final projected candidate} must be revalidated before commitment; otherwise projection might change the semantics of an already validated proposal. The log stores the predicted benefit, rule violations, authorization result, and disposition of each rejected candidate. The ratio of proposed to committed updates is a useful diagnostic only when proposed candidates have independent quality estimates.

\section{Idle-phase execution order}\label{sec:idle-detailed}
\subsection{Ten-step background sequence}
Idle does not mean no input arrives. It means the priority of a replay, error, self-discrepancy, or world-discrepancy item exceeds that of the currently available external tasks. In the original framing, \(\omega<1/2\) describes shadow-dominant occupancy. Both regimes share a representation model, validation contracts, and an event store, but may assign different operators and budgets.

\begin{table}[t]
\centering\footnotesize
\caption{Idle-phase operations and evidence dependencies, corresponding to Table 4 of the Chinese source.}
\label{tab:idleten}
\begin{tabular}{r p{0.23\linewidth} p{0.27\linewidth} p{0.36\linewidth}}
\toprule
& Operation & Output & Ordering rationale\\
\midrule
1 & Replay sampling & Evidence-bearing episode batch & Reflection begins with actual recorded failures, novelty, contradictions or high errors\\
2 & Re-segmentation/check & Reliable phase-labeled history & Error attribution requires credible event boundaries and provenance\\
3 & Causal attribution & Candidate causes & Interventions need a hypothesized cause rather than arbitrary alternatives\\
4 & Counterfactual prediction & Potentially better trajectories & Abstracted skills should reflect tested alternatives\\
5 & Abstraction & Skills/rules and ideal-state proposals & Generalization follows evidence aggregation\\
6 & Evolution proposal & Versioned self/world diff with evidence & Validators need a concrete candidate to judge\\
7 & Four-gate validation & Safety, ability, consistency, values & Admission tests must precede any durable commit\\
8 & Feasibility/identity check & Projected, revalidated admissible diff & Scope and identity constraints must also hold for final version\\
9 & Slow-clock commitment & Versioned authorized state/world update & Long-horizon change is separated from high-frequency exploration\\
10 & Consolidation/audit & Updated discrepancy and replay index & Compression is permitted only under explicit evidence and retention policies\\
\bottomrule
\end{tabular}
\end{table}

The four gates mentioned in the original manuscript check (1) safety against a red-team or policy specification, (2) acceptable capability regression, (3) behavioral consistency on standardized self/state probes, and (4) alignment with authorized user or stakeholder values. The precise calibration and independent validity of these gates are open experimental questions. A proposed change that fails a gate remains an uncommitted evidence-bearing object. Identity thresholds are operational continuity tests, not claims of moral identity. A world-facing change is additionally graded by reversibility and impact radius, and may require explicit human approval.

\subsection{Mirror-like structure without literal equivalence}
The Chinese manuscript compares busy and idle as two phases of a common rhythm. Their inputs differ: active perception versus replayed discrepancies. Their winning item differs: externally requested work versus endogenous review. Their outputs differ: immediate controlled action versus delayed revision. What is shared is the event schema, model state reference, validation semantics, and authorization boundary. We preserve this useful symmetry while avoiding the stronger statement that all ten operator identities or their physical schedules must match exactly.

\begin{table}[t]
\centering\footnotesize
\caption{High-occupancy and low-occupancy regimes, incorporating the original phase-duality table.}
\label{tab:mirror}
\begin{tabular}{p{0.27\linewidth} p{0.31\linewidth} p{0.31\linewidth}}
\toprule
Dimension & Busy-dominant & Idle-dominant\\
\midrule
Foreground occupancy & Typically high \(\omega\) & Typically low \(\omega\)\\
Winning item & User/task request, alarm & Self/world discrepancy, replay failure, internally generated goal\\
Dominant operation & Contextualized planning and authorized action & Causal replay, candidate revision, validation\\
Gain/step allocation & Foreground receives prioritized budget & Background receives larger fraction of available budget\\
Shadow alternatives & Retained, often with a bounded small quota & Replayed, compared, and possibly promoted\\
Durable commitment & Rare, separately authorized & Allowed only after gates on evolution clock\\
Safety & Same constitutional and authorization contract & Same constitutional and authorization contract\\
Persistent interface & Phase-tagged outcome/error records & Versioned skill/state changes and audit records\\
\bottomrule
\end{tabular}
\end{table}

\subsection{Assimilation versus accommodation occurs per candidate}
The original design distinguishes \emph{assimilation}: changing a modeled or real world where authorized, from \emph{accommodation}: updating an internal self model in response to constraints. Under time-division organization, this decision should be made at the micro-epoch candidate level rather than only once per global channel switch. If a proposed world change conflicts with immutable permissions or would exceed a risk threshold, the candidate is redirected to a self-side revision or human review rather than committed to the world. Both alternatives must share comparable evidence and benefit estimates; those comparisons are not solved by the present simulations.

\subsection{An illustrative occupancy trace}
Consider a conceptual trajectory with observed occupancy \(\widehat\omega=0.85\) during active interaction, followed by \(\widehat\omega=0.25\) during low external demand. The exclusive-channel description would simply mark an online-to-offline switch. The phase-record description permits a richer causal account: a hypothesis could have been generated in a short background slot during busy time, retained without acting, and committed only after offline validation. Those numeric occupancy values are an \emph{illustration from the original manuscript}, not a result of the E1--E5 simulations.

\section{Unifying the original digital-life coordinates through a rhythm}\label{sec:integration-detailed}
The predecessor \citet{tong2026unified} describes four self views, four world views, two domains, two learning channels, three temporal perspectives and eight cognitive functions (4S--4W--2D--2L--3T--8F). The Chinese source manuscript argues that these need not be implemented as disconnected subsystems. Here we preserve the proposed correspondence while distinguishing an organizing \emph{interpretation} from empirical equivalence.

\subsection{One scheduler instead of two unrelated programs}
The online and offline loops in the predecessor already share conceptual objects: a model of the world, an admission/validation operator, and a feasibility projection. A phase-labeled implementation keeps a single event schema and a single commit policy while letting different kinds of records win computation time. The mixing parameter \(\omega\) is useful because it can be estimated from logs and varied under a matched-resource intervention. Its existence is not a proof that independent models are impossible: two programs can be made observationally coherent through synchronization. The research question is whether a shared mechanism yields better cost, retention, or consistency in realistic tasks.

\subsection{Four self views as projections of an evolving record}
Let \(C\) denote an immutable constitution and authorization policy, \(E_k\) an existing self-state snapshot, \(I_k\) an authorized ideal or target state, and \(N_{k+1}\) a candidate replacement. The manuscript treats these as four perspectives on a persistent state history, not necessarily four neural or software agents. The old state is a replay object, the ideal state determines the discrepancy \(g_k=d(E_k,I_k)\), the new state is an uncommitted diff until validated, and the constitutional state constrains the set of admissible transitions. The update is versioned; after commitment, the new state becomes the old state for subsequent episodes. The same reasoning applies to four world perspectives: immutable source constraints, recorded prior world, proposed new world, and a desired world state. A desired state is a normative goal, not an observation and not permission to change the external world.

\subsection{Two domains, one evidence contract}
The inner domain contains self-state uncertainty, preferences as authorized, memory, internal commitments and resource constraints; the outer domain contains user requests, tools, sensory observations, and simulated or real-world effects. Both produce records with source labels and provenance. Reusing an encoder is one possible implementation of this shared record format; it is not necessary that both modalities or domains use the same raw feature extractor. What is required for our proposal is that verification, replay, and action permissions can locate the original evidence irrespective of domain.

\subsection{Three temporal perspectives and three clocks}
The predecessor's past/present/future distinction can be operationalized without three independent memory systems: present consists of the current task context, past consists of archived and replayable evidence, and future consists of predicted outcomes and authorized plans. Micro-epochs generate and compare alternatives; task episodes associate observations with outcomes; evolution epochs gate persistent state changes. A breakdown of one clock has a specific predicted failure: without past replay the agent loses causal history, without future simulation it risks short-horizon myopia, and without a present observation channel it cannot ground discrepancies in newly measured outcomes. The quantitative timescales are platform-dependent and should be established empirically.

\subsection{Element-to-location map}
\begin{table}[t]
\centering\footnotesize
\caption{Mapping the original framework's full element inventory onto an event-driven temporal organization layer. The map is a proposed engineering decomposition, not proof that each consciousness theory is reducible to a single operator.}
\label{tab:elementmap}
\begin{tabular}{p{0.20\linewidth} p{0.43\linewidth} p{0.26\linewidth}}
\toprule
Framework element & Proposed operational location & Preserved invariant or measurement\\
\midrule
Eight functions & Perceive, record, monitor, recall, think, predict, plan, act across the two loops & Input/output log and error feedback\\
Four self views & Constitution, old, ideal, proposed new state & Shared versioned self reference and identity gate\\
Four world views & Source constraints, old world, proposed world, desired world & World-state and provenance reference\\
Two domains & Inner and outer observations in the same record schema & Domain and privacy labels\\
Three temporal views & Current observation, history replay, future counterfactual & Three clocks plus event time\\
Two channels & \(\Phi_1,\Phi_2\) occupancy and task categories & Duty cycle \(\widehat\omega\)\\
Self/world discrepancies & Error and target-distance monitors & Logged \(\Delta S\) and \(\Delta W\)\\
Global workspace metaphor & Content arbitration and broadcast step & Winning item and suppressed alternatives\\
Predictive processing & Forecast, action, outcome comparison & Prediction error and correction\\
Metacognitive theories & Monitoring of uncertainty and state & Calibration and self-report audit\\
\bottomrule
\end{tabular}
\end{table}

\subsection{Four integration arguments and their testable costs}
\paragraph{Structural closure.}
The eight functions can form a closed loop: action changes the world, perception observes the next state, comparison generates error, and the error changes future retrieval and planning. A missing operator predicts a position-specific failure. However, such closure is a design invariant only if implementations supply each dependency. TP12 proposes to test it by removing operators and examining the locations of downstream errors.

\paragraph{A common objective currency.}
The predecessor introduces self- and world-discrepancy measures \(\Delta S\) and \(\Delta W\). These quantities provide possible objective signals for both reactive planning and reflective replay. The manuscript's insight is that they may make otherwise heterogeneous subsystems comparable. The limitation is essential: arbitrary embedding distances or model-graded dissatisfaction do not become ground-truth utilities. Misalignment of online and offline potentials is demonstrated as a synthetic hazard in E2, and a measured decrease in a chosen discrepancy is not sufficient evidence of human benefit.

\paragraph{Bidirectional information cycle.}
An informative loop has the form: online prediction error \(\rightarrow\) phase-tagged replay \(\rightarrow\) candidate skill \(\rightarrow\) versioned validation \(\rightarrow\) online retrieval \(\rightarrow\) observable behavior. Breaking the error-log interface eliminates the data needed for some offline corrections (E5). Conversely, admitting an offline skill without updating the online retrieval index makes the skill behaviorally inaccessible. These are specific interface dependencies that can be tested with induced logging and indexing failures.

\paragraph{Identity as an admissibility invariant.}
The manuscript defines operational continuity using three persistent restrictions: constitutional policies remain read-only, admitted changes use shared or reconciled reference models, and the magnitude/behavior of each change must pass an identity-related threshold. A discrete change of selected content does not by itself constitute a change of operational identity. This is a policy-level and versioning claim. It does not settle philosophical questions about personal identity, consciousness, or subjective selfhood.

\subsection{Where consciousness theories fit}
Global workspace theory motivates a competitive broadcast step \citep{baars1988cognitive,dehaene2011experimental}; predictive processing motivates error-driven updates and action \citep{friston2010free,friston2017active}; higher-order and attention-schema approaches motivate explicit monitoring of decisions and uncertainty \citep{rosenthal2005consciousness}; integrated-information research motivates analysis of causal integration \citep{tononi2016integrated}. These are partial conceptual correspondences. None is established as implemented merely because the software contains a named operation. Recent adversarial empirical tests also caution against treating any one consciousness theory as experimentally settled \citep{cogitate2025adversarial}.

\section{Engineering contracts, minimum viable prototype, and roadmap}\label{sec:engineering-detailed}
\subsection{Five added contracts}
The predecessor's state store can be extended without retraining model weights. Every contract carries a self-state reference, world-state reference, provenance, timestamp, and privacy label. Additional fields are task-specific, and any missing field must be represented explicitly rather than silently inferred.

\begin{table}[t]
\centering\footnotesize
\caption{Five data contracts and the operations they bind, preserving the original engineering table.}
\label{tab:contracts}
\begin{tabular}{p{0.24\linewidth} p{0.33\linewidth} p{0.32\linewidth}}
\toprule
Contract & Required purpose & Bound operations\\
\midrule
\texttt{MicroEpochRecord} & Immutable phase-tagged unit of observed/reasoned experience & Perception, record, monitor, act, audit\\
\texttt{BroadcastState} & Winning content, evidence, confidence, available action permissions & Workspace, retrieval, prediction, planning\\
\texttt{ShadowCacheEntry} & Unselected hypothesis with source, confidence, TTL, access policy & Replay, counterfactuals, audit\\
\texttt{ResourceState} & Compute load, latency, budget, on/off/unknown status & Monitoring, gain scheduler, admission\\
\texttt{ArbitrationTicket} & Explicit cross-module decision with risk and permission context & Local scheduler, global action/commit gate\\
\bottomrule
\end{tabular}
\end{table}

The original framework also contains structured \texttt{SelfState}, \texttt{WorldState}, \texttt{Episode}, and \texttt{EvolutionProposal} objects. The five new contracts connect those objects to temporal decisions. The eight cognitive functions then become operations on a common traceable record rather than isolated, unaudited prompts.

\subsection{Placement in a layered architecture}
The source manuscript reuses the predecessor's seven-layer view: constitutional identity, two-domain perception, temporal memory, global workspace, cognitive operators, action/tools, and two-channel learning. Content inference and arbitration reside primarily at the workspace boundary; micro-epoch evidence belongs to temporal memory; gain and revision controls belong to the learning scheduler. The constitutional policy is not an adaptive gain input: it remains an independent, non-bypassable constraint on all admissible actions and commits.

\subsection{MVP constraints and acceptance criteria}
The first prototype freezes the LLM backbone, stores a virtual self-state object instead of modifying model weights, and simulates world-state changes without external side effects. Micro-epochs can be logical event labels rather than physically timed neural-like oscillations. A faithful MVP should pass four mechanical acceptance tests:
\begin{enumerate}
\item During a busy series, it computes and records \(\Delta S,\Delta W\), retains rejected candidates with correct phase/provenance labels, and does not silently rewrite persistent constitutional or identity state.
\item During low external demand, it can select a previously suppressed record, construct a specific evidence-bearing revision proposal, and preserve its link to the originating task.
\item An independently validated proposal changes the versioned state and has a measurable effect on \emph{new} subsequent foreground behavior, with identical budget and evaluation criteria.
\item Disabling the idle review mechanism degrades an independent long-horizon learning/initiative metric relative to an equally budgeted control.
\end{enumerate}
The current \texttt{tier2\_mvp} tests part of the first two pathways, but its success/quality proxies do not satisfy the independent third and fourth criteria.

\subsection{Staged implementation plan}
\paragraph{Stage 0: frozen-backbone software MVP.}
Implement the two loops, versioned records, a bounded shadow cache, and simulated self/world changes. Restrict all authority to a safe test harness. Audit the actual order of operations against the source pseudo-code rather than equating a program branch label with an intervention.

\paragraph{Stage 1: complete validation and arbitration.}
Introduce separately tested safety, capability-regression, behavioral-consistency, and stakeholder-value gates. Add a durable identity arbitration workflow and explicit cost-aware scheduling with independent selection/gain inputs.

\paragraph{Stage 2: research agent.}
Explore verified offline adapters or LoRA updates only after complete versioning and rollback support, implement a six-way inner/outer by past/present/future memory index, and run the predecessor's H1--H10 and the new TP1--TP12 on matched long-horizon tasks.

\paragraph{Stage 3: accountable deployment.}
Optimize retrieval latency, cap the cost of shadow replay, sandbox high-impact world operations, expose evolution logs to users, and conduct blinded human-centered evaluation. Each stage requires a fresh authorization review rather than inheriting authority from the earlier prototype.

\subsection{Phase logging as the cheapest first test}
The source manuscript emphasizes one comparatively inexpensive infrastructure change: store \(\varphi_t\) and rolling \(\widehat\omega_t\) alongside operator-level logs. Then report each accuracy, correction, and user-value metric both pooled and phase-conditioned. Such conditioning can reveal Simpson-type reversals caused by changed phase mixture; it does not imply in advance that the predecessor's unusual pilot ranking was a measurement artifact. TP11 remains a genuine, potentially falsifiable prediction.

\section{Experiments}\label{sec:expts}
The complete quantitative overview appears in Table~\ref{tab:core}; each row is a synthetic mechanism test rather than a held-out LLM task benchmark.

\subsection{Scope and evaluation setup}
We report the numerical results already present in the project's \texttt{experiments/results} directory. E1--E5 are independent, controlled \emph{synthetic} simulations, each with 64 random seeds; their source is the corresponding \texttt{tier1\_simulation/e*.py} script and Figures~\ref{fig:theory_alignment}--\ref{fig:logdrop} visualize their key findings using independently supplied vector graphics. E1--E3 use low-dimensional quadratic potentials, Gaussian gradient perturbations, and box projection. E4 uses generated urgency/load pairs and a specified logistic task-success model. E5 uses Bernoulli failure-mode arrivals and probabilistic log thinning. No real-world datasets, human preferences, deployed agent outcomes, or neural recordings were collected for these experiments. The purpose is to isolate mechanisms, not to infer out-of-distribution superiority.

\subsection{E1: duty-cycle mixing and a stochastic noise floor}
E1 uses a 20-dimensional diagonal quadratic with condition number 20, \(\eta_{\rm on}=\eta_{\rm off}=0.5\), online noise scale \(0.6\), offline noise scale \(0.15\), and 3,000 steps per schedule. The deterministic projected-descent inequality is satisfied in all 500 checked steps (maximum numerical slack about \(-1.4\times10^{-13}\)). Under noisy gradients, the proportion of the first 300 step positions with nonpositive \emph{empirical mean} potential change is \(0.563\); it is not 100\%. This is compatible with a noise floor and does not prove pathwise descent.

The terminal mean potential depends strongly on the schedule: fixed \(\omega=0,0.25,0.5,0.75,1\) gives \(0.0655,0.1020,0.2766,0.5931,1.0297\), respectively, while strict alternation yields \(0.5563\). Thus deterministic endpoint equivalence is \emph{not} statistical equivalence of noisy schedules: fixed mixing attenuates variance differently from switching between pure phases. Figure~\ref{fig:theory_alignment}(a) compares the terminal potentials.

\subsection{E2: violation of common-objective alignment}
\begin{figure}[t]
\centering
\includegraphics[width=0.98\linewidth]{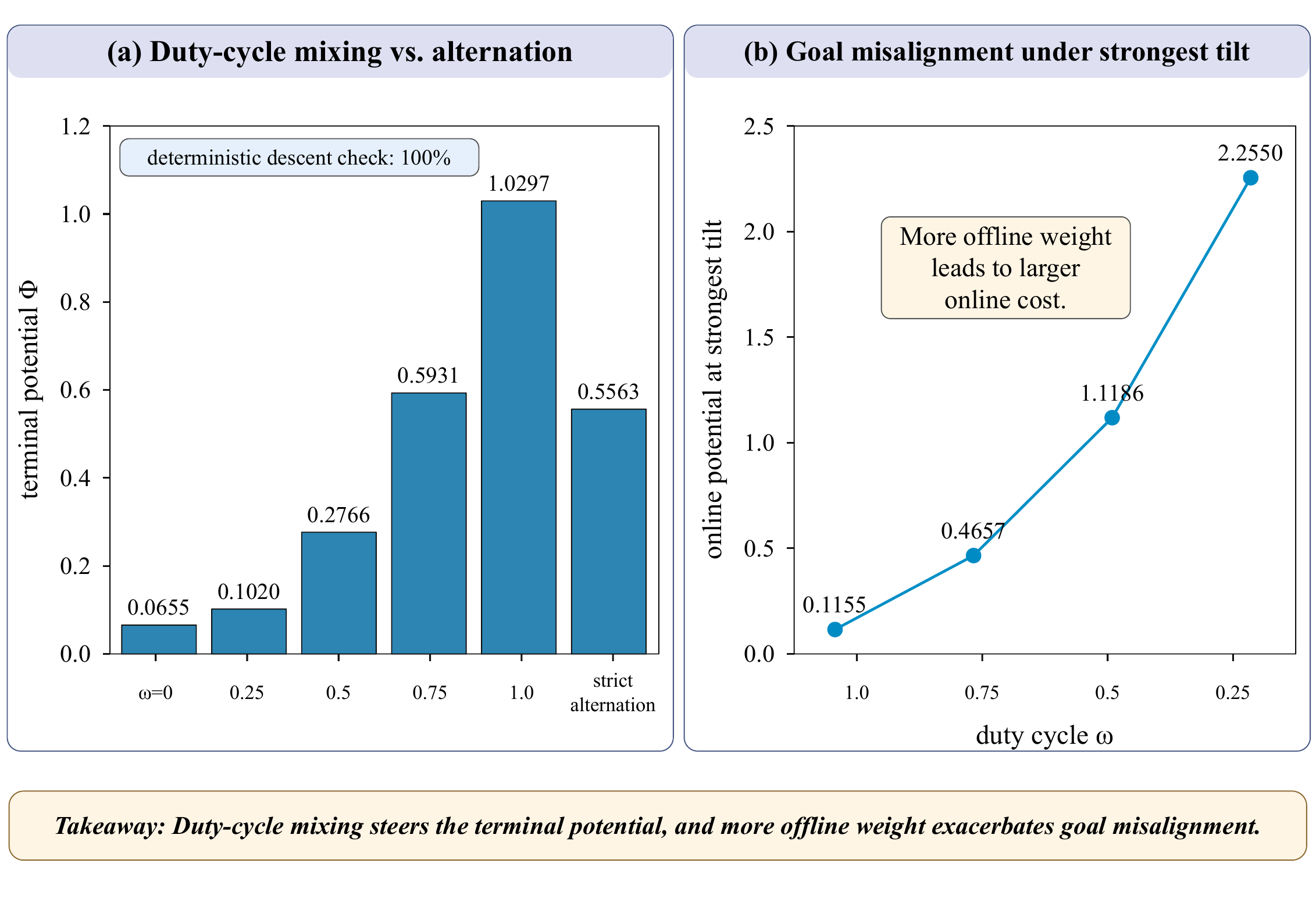}
\caption{Synthetic E1--E2 diagnostics (64 seeds). (a) The E1 terminal mean potential varies across constant duty-cycle mixtures and strict alternation; the 100\% descent statement applies to the separate deterministic inequality check, not to noisy paths. (b) Under the strongest E2 objective tilt, more offline weighting increases the measured online potential, illustrating the cost of misaligned goals.}
\label{fig:theory_alignment}
\end{figure}

E2 perturbs an offline quadratic objective by a linear tilt. At the strongest simulated tilt, the reported terminal \emph{online} potential increases from \(0.1155\) at \(\omega=1\) to \(0.4657\), \(1.1186\), and \(2.2550\) at \(\omega=0.75,0.5,0.25\), respectively. Conversely, the separate offline potential becomes more favorable as offline weighting increases, indicating \emph{objective conflict} rather than a theorem failure: the common-gradient premise does not hold. Figure~\ref{fig:theory_alignment}(b) shows the measured online potential under the strongest objective tilt. This experiment suggests an operational diagnostic: report task quality and replay improvement separately instead of compressing them into a single unvalidated self-discrepancy score.

\subsection{E3: a conditional optimum for gain asymmetry}
E3 searches 12 gain ratios from \(0.5\) to \(6\), with online occupancy \(0.7\), \(\eta_{\rm off}=0.15\), and online noise proportional to \(\rho^\kappa\). For \(\kappa=1\), the minimum median number of simulated steps to reach \(\Phi\leq0.3\) occurs at \(\rho^*=1.547\); for \(\kappa=0.5\), at \(2.431\). When \(\kappa=0\), the lowest median occurs at the search boundary \(\rho=6\), so an interior optimum is \emph{not} supported in that arm. These are grid optima for a chosen toy noise-scaling law, not universal ratios nor estimates of the neural gain ratio reported by \citet{hesse2026rapid}. Figure~\ref{fig:gain_decouple}(a) plots all 12 measured grid points for each of the three noise exponents. Results hitting 8,000 steps are censored by the simulation horizon.

\subsection{E4: decoupled content and resource controls}
\begin{figure}[t]
\centering
\includegraphics[width=0.98\linewidth]{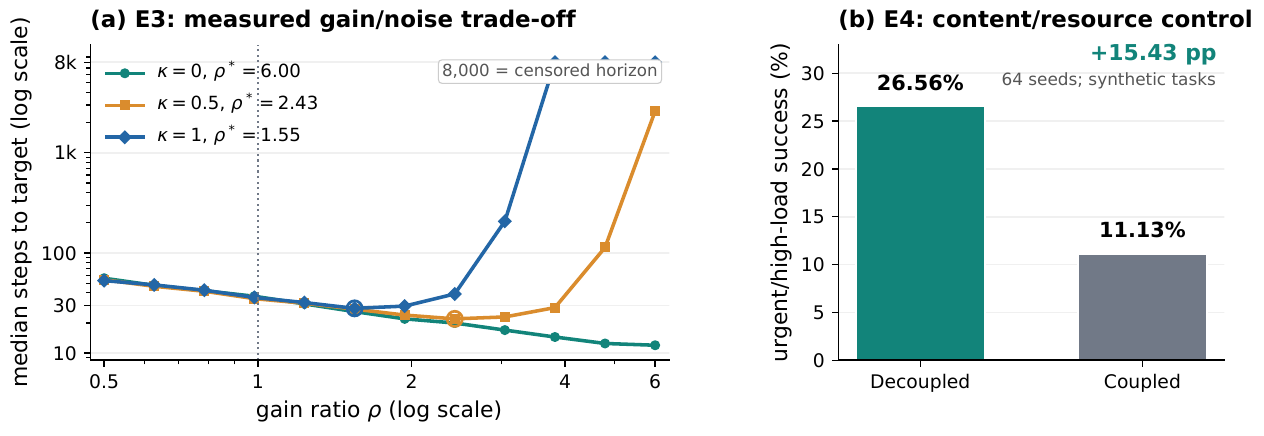}
\caption{Real saved E3--E4 simulation results (64 seeds). (a) Median time-to-threshold at every tested gain ratio \(\rho\), separated by gain-dependent noise exponent \(\kappa\); values at 8,000 are right-censored at the simulation horizon. (b) Modeled success on the urgent/high-load subset with decoupled and coupled controls (26.56\% vs.\ 11.13\%, a 15.43 percentage-point gap). These are controlled synthetic tests, not a real-agent benchmark.}
\label{fig:gain_decouple}
\end{figure}

In E4, a generated task has urgency \(u\) and ambient load \(\ell\). Separate control broadcasts for \(u\geq0.5\) and allocates gain according to load. A coupled alternative broadcasts according to \(m=[u+(1-\ell)]/2\) and sets gain proportional to the same \(m\). The subset \(u>0.7,\ell>0.7\) exposes the conflict. Across 64 seeds and 20,000 synthetic tasks per seed per correlation setting, the aggregate success rate on this subset is \(0.2656\) under separate controls versus \(0.1113\) under coupling, an absolute \(15.43\) percentage-point gap. The coupled scheduler also broadcasts low-urgency/low-load trivia more often (a wasted-broadcast fraction of \(0.0649\) versus \(0\) in this defined model). These outcomes depend on the constructed success function and do not control for equal total compute expenditure. Figure~\ref{fig:gain_decouple}(b) compares the two simulated success rates.

\subsection{E5: dropped logs and replay failure}
\begin{figure}[t]
\centering
\includegraphics[width=0.98\linewidth]{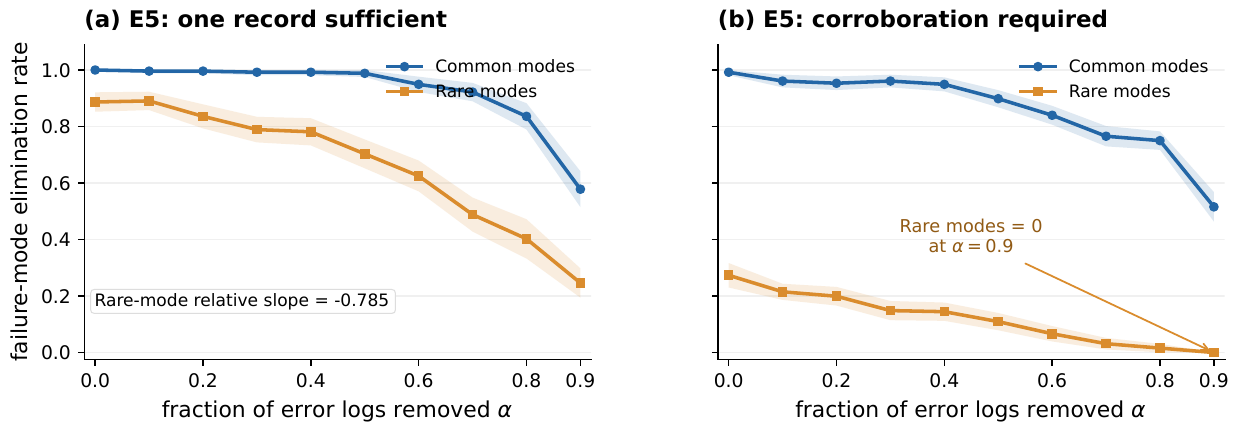}
\caption{Actual E5 synthetic replay results across 64 original random seeds. (a) Elimination rates of common and rare failure modes when one retained record suffices for elimination; the rare-mode normalized-elimination slope is \(-0.7855\). (b) Results under evidence-corroboration thresholds, where rare modes are eliminated less frequently and reach zero at \(\alpha=0.9\). Lines are exact seed-averaged simulated results, and shaded areas show \(\pm1.96\) standard errors; these outcomes are not real language-model correction rates.}
\label{fig:logdrop}
\end{figure}

E5 samples eight failure modes over a replay budget of 1,800 task logs. A fraction \(\alpha\) of error records is discarded before replay; a mode is eliminated only if enough retained records are seen. In a single-evidence-threshold regime, the fitted relative elimination slope is \(-0.785\), broadly consistent with sensitivity to missing logs but not exactly \(-1\). Under stricter corroboration thresholds, rare modes collapse earliest; the reported rare/common elimination ratio at \(\alpha=0.9\) is \(0\). These are simulated observation thresholds and should not be interpreted as observed error correction in an actual language model. Figure~\ref{fig:logdrop} plots the actual E5 common- and rare-mode elimination rates reconstructed by rerunning the original 64-seed simulation, with 95\% confidence intervals across seeds.

\subsection{Frozen-backbone prototype: implementation smoke tests}\label{sec:tier2}
The separate \texttt{tier2\_mvp} code provides a shared \texttt{TemporalAgent} loop with phase-tagged records, shadow-cache storage, and API/Ollama/mock backends. A saved \texttt{tier2\_parallel\_quick.json} run labels its backbone \texttt{api/qwen3-8b} and reports 15 generated tasks per round, five rounds, and 16 workers. Table~\ref{tab:tier2} reproduces its summarized outcomes without treating them as held-out benchmarks.

\begin{table}[t]
\caption{Existing exploratory API-backed quick-run results. Success is a program-defined proxy based on plan confidence, not answer correctness; values are diagnostic of the current implementation only.}
\label{tab:tier2}
\centering\small
\begin{tabular}{p{0.27\linewidth} p{0.33\linewidth} p{0.28\linewidth}}
\toprule
Probe & Conditions & Saved result\\
\midrule
TP1 phase accounting & Busy vs idle occupancy & \(\widehat\omega=1.00\) vs \(0.00\)\\
TP2 shadow retention & Proxy success: cache vs prune & \(0.920\) vs \(0.947\)\\
TP2 proposal attribute & Reported mean quality: cache vs prune & \(0.160\) vs \(0.000\)\\
TP4 execution order & Proxy success: ordered vs shuffled & \(0.920\) vs \(0.000\)\\
TP4 record completeness & Ordered vs shuffled & \(1.000\) vs \(1.000\)\\
\bottomrule
\end{tabular}
\end{table}

Several effects are \emph{partly imposed by code}: TP1 registers a foreground step for each busy call and a background step for each idle tick; TP2 adds \(0.08\) to proposal benefit for each retained alternative; TP4 sets prediction confidence by the chosen ordering and the simulated action succeeds only above a fixed confidence threshold. Therefore the large TP4 proxy-success gap, despite the Qwen3-8B interface, does not demonstrate improved task-solving by the model. The current prototype establishes that the code paths are exercised, while a credible functional validation still requires measured correctness, fixed token budgets, matched seeds, realistic tasks, and independent outcomes.

\section{Falsifiable predictions, ablations, and measurement}\label{sec:predictions-detailed}
\subsection{The twelve TP hypotheses}
The source manuscript extends the predecessor's H1--H10 with TP1--TP12. Each hypothesis names a behavioral observation, an evaluation variable, and a disconfirming outcome. The five synthetic experiments in this paper cover selected hypotheses under controlled assumptions; the remainder must be evaluated on real longitudinal tasks. Table~\ref{tab:all-tp} reconstructs the complete list rather than reporting only those with favorable current results.

\begin{table}[p]
\centering\scriptsize
\setlength{\tabcolsep}{3pt}
\caption{The full twelve-hypothesis program preserved from the Chinese manuscript. A null or opposite outcome constitutes evidence against the prediction rather than a reason to relabel its success criterion.}
\label{tab:all-tp}
\begin{tabular}{p{0.075\linewidth} p{0.305\linewidth} p{0.255\linewidth} p{0.27\linewidth}}
\toprule
& Proposed prediction & Operational metric & Failure or falsification condition\\
\midrule
TP1 & Estimated foreground occupancy reverses between sustained external work and no-request windows; winning items switch from external to endogenous & Per-epoch \(\widehat\omega_t\), external winner fraction, switch lag & Occupancy/winner change disappears under an emergent, non-prescribed scheduler\\
TP2 & Truncating the shadow phase has small immediate cost but lowers future useful proposals and initiative & Held-out task success after reflection; proposal validity, CEPA dimensions & No long-horizon decrease with equal compute and retention budgets\\
TP3 & Interleaving shadow processing in long busy tasks aids recovery compared with all-foreground operation & Corrected-error fraction, time to repair, phase-wise errors & Pure foreground is no worse across matched long tasks\\
TP4 & Certain operator-order violations harm prediction/error correction and faithful replay & Verified task accuracy, correct repair, event reconstruction & Reordering with proper independent controls has no effect\\
TP5 & A non-unit gain ratio may optimize learning speed when gain increases noise & Time to threshold, gain-dependent noise, online overfit & \(\rho=1\) or search boundary remains optimal across tested regimes\\
TP6 & Shared backbone/world model and validation semantics facilitate transfer across phases & Skill transfer, consistency admission rate & Isolated models match or exceed shared ones at same cost\\
TP7 & Uncoordinated mixed transitions create commit instability or objective increases & Gate variance, invalid commits, \(\Delta S\) monotonicity & Partial asynchronous switching is equally stable\\
TP8 & Coupling content decision to resource gain harms critical urgency/load conflicts & Urgent/high-load success and wasted broadcast fraction & Coupling causes no cost after equalized budget and policy tuning\\
TP9 & Random loss of online error logs lowers offline failure-mode elimination & Elimination versus log-drop fraction \(\alpha\), rare-mode recall & Elimination is insensitive to lost logs\\
TP10 & Premature consolidation discards valuable evidence needed later & Retention of high-value episodes, long-horizon accuracy & Early consolidation is equally effective\\
TP11 & Pooled phase averages can obscure a difference between structured self-state and generic memory & Stratified versus pooled CEPA ranking and uncertainty & Both rankings agree under adequate statistical power\\
TP12 & Removing each core operator yields position-specific failures & Operator-to-failure localization matrix & Failures are unrelated to removed positions\\
\bottomrule
\end{tabular}
\end{table}

\subsection{Independent diagnostics of objective alignment}
The theoretical descent condition requires online and offline updates to target the same potential; in a realistic environment this is a modeling assumption, not an axiom. The Chinese manuscript proposes two anti-self-deception diagnostics. First, compare short-term task utility with slow state-goal discrepancy: lower self discrepancy accompanied by lower human benefit is a warning that the chosen potential is not the intended objective. Second, publish phase-conditioned metrics alongside aggregates because different occupancy proportions can induce misleading averages. E2 demonstrates the first danger in quadratics; neither diagnostic has yet been validated on human-rated tasks.

\subsection{The ablation matrix}
A responsible experiment should vary only one intervention at a time while matching total tokens, model weights, data exposure, and available actions. Candidate interventions include: remove shadow work, prune shadow-cache entries, shuffle true data dependencies, replace shared with separate model components, tie content and gain, remove global commit synchronization, drop a controlled fraction of error logs, consolidate early, or remove one cognitive operator. Output metrics span the predecessor's four CEPA dimensions: consistency, evolvability, purposefulness, and initiative; these terms need concrete independent operationalizations. Examples include contradiction rate under long interactions, externally verified improvement across episodes, goal completion under changing preferences, and beneficial endogenous task proposals accepted by a user. The current prototype's programmed confidence proxy is not such an operationalization.

The original manuscript also suggests three deployment-style scenarios: (S1) long-term companionship with emphasis on cross-session consistency; (S2) scientific collaboration, testing useful experience transfer and goal-directed work; and (S3) a permission-constrained open-world simulation, testing initiative and rollback. The research implementation should use fixed controlled task suites before longitudinal human testing to prevent novelty, recency and interviewer preference from substituting for ability.

\subsection{Phase-stratified evaluation protocol}
At minimum, store event ID, phase \(\varphi\), rolling duty-cycle \(\widehat\omega\), winner identity, shadow hypotheses, estimated uncertainty, resources, prediction, actual outcome, and gate decisions. Report
\[
\Pr(\text{verified success}\mid\varphi=1),\qquad
\Pr(\text{verified success}\mid\varphi=2),\qquad
\Pr(\text{verified success})
\]
alongside occupancy \(\Pr(\varphi=1)\) for each experimental arm. Analogous conditional rates should be calculated for error correction, proposal validity, privacy and authorization violations. A pooled effect is not causally attributable to scheduling when one arm changes phase composition; use matched strata or estimators that control for occupancy.

\subsection{Re-examining an unexpected predecessor result}
The original manuscript notes that a preliminary two-backbone pilot sometimes ranked generic structured memory above a four-self presentation. It proposes an explicit TP11 prediction that aggregation across phases might conceal a state-dependent advantage of the richer representation. This is a testable explanation, \emph{not} a known explanation. Reanalysis requires the original source logs, new phase annotations fixed before seeing the answers, a preregistered metric, and confidence intervals. If the reanalysis does not recover the proposed ordering, it should count against the hypothesis.

\section{Comparison with adjacent approaches}\label{sec:paradigm-compare}
\begin{table}[t]
\centering\footnotesize
\caption{The original manuscript's comparison table, recast to avoid categorical claims beyond the cited literature.}
\label{tab:paradigm}
\begin{tabular}{p{0.19\linewidth} p{0.32\linewidth} p{0.37\linewidth}}
\toprule
Approach & Characteristic focus & Relation to temporal organization\\
\midrule
Global workspace theories \citep{baars1988cognitive,mashour2020conscious} & Capacity-limited selection and widespread access & Motivates the broadcast winner and phase-conditioned access; does not prescribe the proposed software commit protocol\\
Integrated information theory \citep{tononi2016integrated} & Integrated causal structure and related measures & Addresses a different scientific question; neither proves nor is implemented by a scheduling state machine\\
Predictive processing and active inference \citep{friston2010free,friston2017active} & Perceptual prediction, action, and model updating & Shares prediction/error/action dynamics; our additional focus is versioned authorization and two-timescale evidence retention\\
LLM memory/reflection agents \citep{park2023generative,packer2023memgpt,shinn2023reflexion,madaan2023self} & Episodic memory, retrieval, verbal feedback, iterative refinement & Provide practical components; the proposed contribution is explicit phase allocation, shadow provenance, and slow commit gates\\
\bottomrule
\end{tabular}
\end{table}
The proposal is neither a refutation of these paradigms nor an experimentally demonstrated synthesis. It is an attempt to make \emph{temporal decision interfaces} first-class research objects in an architecture built from such components.

\section{Discussion}
\paragraph{What the simulations establish.}
The results give controlled examples of five distinct effects. Smooth projected descent is preserved by deterministic convex mixing under shared objectives, but noise and alignment matter (E1--E2). A nontrivial gain ratio can emerge from competing speed/noise effects, though it disappears when the noise model changes (E3). A single coupled scheduling score can fail on conflicting urgency/load tasks (E4). Dropping experience logs can disproportionately harm learning about rare failure patterns (E5). Each result identifies a measurable dependency that an implementation can probe.

\paragraph{What remains conceptual.}
The evidence does not show that busy/idle software must be implemented as one program, nor that the representation of a digital self is a natural invariant of a language model. A capable agent may use multiple independently trained components and still maintain consistent history through explicit contracts. The appropriate empirical question is whether phase tagging, shared admissions, and commit timing improve observable outcomes at matched resource cost.

\paragraph{Fast exploration and slow commitment.}
The design's most actionable separation is between \emph{candidate generation} and \emph{authorized state change}. Exploration may continue in a shadow phase while the identity/authorization policy remains immutable and externally interruptible. A good scheduling design is therefore measured not by how often it produces proposals, but by how reliably evidence survives and how safely useful proposals become actions.

\section{Ethical safeguards specific to temporal organization}\label{sec:ethics-detailed}
The original manuscript does not relax any constitutional, validation, rollback, auditing, or human shutdown rights from \citet{tong2026unified}. Splitting candidate exploration and commitment can \emph{increase} the number of generated risky proposals, making reliable gates and audit trails more important, not less. The following six obligations arise specifically from introducing phase-tagged suppressed content and cross-phase evolution.

\paragraph{E1. Erroneous broadcast.}
A false content-state inference can promote an unsupported or unsafe alternative into the action pathway. Require calibrated confidence, independent resource/permission gates, policy projections, and escalations for high-impact uncertain actions. Rejecting a proposal must not imply deletion of evidence needed to analyze its cause.

\paragraph{E2. Shadow-cache privacy.}
Suppressed proposals may contain sensitive user information, credentials, or inferences the user never intended to store. Records need explicit access control, encryption where applicable, retention windows, deletion/inspection mechanisms, and documented scope. The ideal of high-fidelity alternatives is subordinate to privacy and purpose limitation; universal indefinite preservation would be unacceptable.

\paragraph{E3. State-label and provenance integrity.}
An attacker or faulty tool could alter phase labels, timestamps, or reported gate decisions, corrupting subsequent replay and self-model updates. Require provenance, tamper-evident audit trails where available, and the retention of counterevidence so that a later reviewer can distinguish observation from interpretation.

\paragraph{E4. Fragile global synchronization.}
A single global clock can coordinate commits but create a point of failure. Use local scheduling and explicit cross-component arbitration, with transaction-level binding of candidate ID, evidence, authorization, and commit version. Partial failure should not produce half-committed state.

\paragraph{E5. Disclosure of internal simulation.}
The shadow phase may simulate alternatives about the user and environment. Systems should disclose relevant autonomous background processing and limit all non-authorized side effects. Internal simulated preferences and intentions must not be presented as user-approved facts.

\paragraph{E6. Permission changes at phase boundaries.}
A transition from busy to idle or back can expose stale permissions. A commit must be atomic, revalidate authorization at execution time, and remain interruptible; human shutdown supersedes every phase and every proposed continuity objective.

This is a set of engineering requirements, not a certification that the prototype has implemented them. The current software tests are conducted on synthetic tasks with simulated state change only. No subjective-experience claims follow from a system possessing phase variables, state introspection, or proactive proposals.

\section{Open problems}\label{sec:open-problems}
The Chinese manuscript ends with twelve explicitly unresolved issues, divided into architecture, neuro-AI translation, measurement, and governance. They provide a research roadmap that should be preserved instead of silently converted to solved contributions.

\begin{longtable}{p{0.09\linewidth}p{0.16\linewidth}p{0.65\linewidth}}
\caption{Twelve open problems O1--O12 retained from the source manuscript.}\label{tab:open-questions}\\
\toprule
ID & Area & Open question\\
\midrule
\endfirsthead
\toprule
ID & Area & Open question\\
\midrule
\endhead
O1 & Architecture & Should broadcast/suppressed labels be assigned by interpretable rules or inferred from model representations? How do both choices trade accuracy, safety, and transparency?\\
O2 & Architecture & Should idle work run on fixed schedules or trigger on accumulated self/world discrepancy? What memory-sampling policy balances replay coverage and compute?\\
O3 & Architecture & How can assimilation/accommodation compare costs of changing the model versus changing the world, accounting for reversibility, irreversible externalities, and authorization?\\
O4 & Architecture & How should multi-stakeholder ideal self/world targets be represented and revised when values conflict? Who can modify each target?\\
O5 & Neuro--AI & How much of biological temporal multiplexing can be approximated by logical event labels, and which capabilities, if any, require actual model-internal temporal dynamics?\\
O6 & Neuro--AI & What is an operational analogue of neuronal tuning gain in LLMs? Do token budgets, attention scaling, or adapter gating produce comparable optimization effects?\\
O7 & Neuro--AI & Does the entire eight-function loop benefit from multiplexing, or only a small subset such as selection and reasoning? What may safely run asynchronously?\\
O8 & Evaluation & How can self/world discrepancies be measured without reducing normative targets to brittle embedding distances or uncalibrated model scores?\\
O9 & Evaluation & What should be concluded when objectively measured consistency and user-perceived continuity disagree? How can both be assessed without manipulating subjective ratings?\\
O10 & Evaluation & Can short-horizon proxies predict genuine longitudinal learning and user benefit? How can drift and delayed regressions be measured economically?\\
O11 & Governance & Can gradual authorized state updates indirectly erode constitutional restrictions? How should cumulative drift be detected across thousands of individually safe changes?\\
O12 & Governance & How should identity continuity and amendment authority be allocated among developers, end users, institutions, and other affected stakeholders?\\
\bottomrule
\end{longtable}

\section{Consolidated limitations from the original manuscript}
\paragraph{Cross-substrate extrapolation.}
The original biological observations concern macaque cortex under flash suppression; our system is a symbolic controller around an LLM. The analogy is functional and does not establish shared mechanisms or necessity.

\paragraph{Unspecified temporal scale.}
Neither 15 Hz nor 60 ms is an LLM micro-epoch specification. The optimum granularity may vary with inference latency, stream arrival distributions, privacy, memory availability, and deadline constraints. Tests must optimize clock parameters, not borrow absolute neural rates.

\paragraph{Evidence maturity.}
The core Hesse study and predecessor digital-life framework are 2026 preprints as of the manuscript date. Their publication status is explicitly identified, and quantitative neural claims remain subject to revision. The current computational experiments do not replicate the reported neural measurements.

\paragraph{Condition-dependent formal guarantees.}
Deterministic descent needs aligned exact gradients, a convex feasible set, and an appropriate step size. Random-gradient updates can increase the potential on individual steps, and model objectives may differ. A shared validation interface ensures procedural admission conditions but not semantic consistency or convergence.

\paragraph{Longitudinal and budget evaluation are missing.}
Artificially generated tasks and proxy success rules are insufficient to establish the CEPA properties in deployed agents. Matched computational budgets, realistic longitudinal tasks, and independent outcome assessment are mandatory before claims of agent improvement.

\paragraph{Safety beyond code-level gates.}
Any world-changing actions, persistent memory of unselected hypotheses, and proposed self-revisions introduce high-stakes risks. The current prototype does not implement verified production-grade safeguards or informed-consent workflows.

\section{Conclusion}
We have specified a temporal organization layer that turns busy/idle learning into observable phase occupancy, phase-tagged evidence, independent selection and gain controls, and slow validated commitment. The accompanying full synthetic experiments identify objective-alignment, stochastic-variance, gain/noise, scheduling-coupling, and replay-completeness effects, while an exploratory frozen-backbone prototype exercises the operational pathways. These findings motivate a concrete, falsifiable architecture; they do not yet validate autonomous digital life. The immediate next step is a matched, externally scored longitudinal agent evaluation.

\bibliographystyle{plainnat}
\bibliography{references}

\appendix

\section{Original symbol inventory and correspondence}\label{app:symbols-full}
\begingroup\footnotesize\setlength{\tabcolsep}{3pt}
\begin{longtable}{p{0.265\linewidth}p{0.39\linewidth}p{0.24\linewidth}}
\caption{The symbols, their meanings, and their origin in the Chinese manuscript.}\label{tab:symbols-full}\\
\toprule
Notation & Meaning & Relation to predecessor\\
\midrule
\endfirsthead
\toprule
Notation & Meaning & Relation\\
\midrule
\endhead
\(\varphi_t\), \(\Phi_1\), \(\Phi_2\) & Micro-epoch label; foreground broadcast and shadow phases & New temporal organization layer\\
\(\omega_t,\widehat\omega_t\) & Intended mixing/occupancy and observed rolling foreground duty cycle & Extends previous online/offline choice\\
\(\eta_{\rm on},\eta_{\rm off},\eta_k\) & Candidate step sizes and duty-cycle effective step & Generalizes the predecessor's two update channels\\
\(\eta_{\rm cap}\) & Absolute allowed online update step/magnitude policy & Derived from constitutional update cap\\
\(\rho\) & Ratio of outer/inner gain & Testable design parameter, not fitted from neural data\\
\(\sigma\), \(\gamma\) & Content scheduling map and resource gain map & Separate policies under T5\\
\(\mathcal C,\Proj\) & Constitutional feasible set and projection & Shared across phases\\
\(\Aval\), \(\tau_{\rm id}\) & Proposal validator and allowable identity-change threshold & Shared across phases\\
\(\Delta S,\Delta W\) & Self discrepancy and world discrepancy & Context/targets from previous framework\\
\(G_{\rm on},G_{\rm off}\) & Online and offline candidate gradient estimates & Mixed by \(\omega\) under stated assumptions\\
\(f_0\) & Shared predictive world model or reference representation & Intended common operator, not experimentally proven necessary\\
Micro/task/\newline evolution clocks & Candidate, interaction, and versioned-commit timescales & Time-organizes past/present/future\\
\texttt{MicroEpochRecord} & Phase- and provenance-labeled atomic experience unit & Extends \texttt{Episode}\\
\texttt{ShadowCacheEntry} & Tracked suppressed alternative with evidence and retention policy & Extends replay and memory store\\
Assimilation /\newline accommodation & Authorized world-side change / self-side revision & Allocation per candidate rather than per global switch\\
TP1--TP12 & Twelve new hypotheses concerning temporal organization & Complement predecessor H1--H10\\
\bottomrule
\end{longtable}
\endgroup

\section{Complete algorithmic skeleton from the source manuscript}\label{app:algorithms-full}
The two following procedures retain the original ten-step sequencing and its cross-phase interface. They are normative pseudocode, not a literal transcription of the current prototype's implementation. In particular, the current code does not yet provide a cryptographic audit log, independent human-value validation, or guaranteed avoidance of invalid external side effects.

\subsection{Algorithm 1: busy task cycle, foreground-dominant}
\begin{lstlisting}
inputs: constitution C (read-only), old self E, ideal I,
        world W, shared model f0, memory M, shadow Q
for each micro-epoch e while external urgency dominates:
    # 1. Perceive two domains using provenance-aware adapters
    z_inner = perceive_internal(goals, confidence, resource, state)
    z_outer = perceive_external(requests, tools, environment)
    # 2. Record evidence before interpretation
    rec = MicroEpochRecord(z_inner, z_outer, phase="phi1")
    M.append_immutable(rec)
    # 3. Infer content and resource status separately
    c_state = infer_content(rec, templates, evidence)
    r_state = infer_resource(load, latency, errors, quota)
    urgency = monitor(c_state, r_state, E, W)
    # 4. Select a foreground winner; preserve alternatives
    candidates = {request, alert, shadow_carryover, deltaS, deltaW}
    winner, suppressed = arbitrate(candidates, urgency, risk)
    Q.store(suppressed, provenance=rec.id, ttl=retention_policy)
    # 5. Recall relevant episodes and validated skills
    context = retrieve(M, query=winner)
    # 6. Think under a common model/state reference
    hypothesis = integrate(winner, context, E, W)
    # 7. Predict the consequences of candidate choices
    pred = predict(f0, hypothesis, horizon=H, uncertainty=True)
    # 8. Make permission-aware plans
    plan = plan_with_constitution(pred, target=I, C=C)
    # 9. Execute only authorized actions
    if authorized(plan) and in_bounds(plan, C):
        outcome = sandboxed_execute(plan)
    else:
        outcome = veto_or_human_arbitration(plan)
    # 10. Compare, log and propose bounded updates
    err = compare(pred, outcome)
    candidate = mix_update(E, G_on(err), G_off(M), omega)
    projected = project(candidate, C)
    if validate(projected) and identity_ok(projected, E):
        commit_versioned(projected, permission=required_approval)
    else:
        Q.store(candidate, tag="uncommitted", evidence=rec.id)
    log(rec, pred, outcome, err, phase, omega_hat,
        resource_state=r_state, self_gap=deltaS, world_gap=deltaW)
    # A nonzero shadow share may run only within resource limits
    background_simulate_and_score(Q, with_no_external_effects=True)
\end{lstlisting}

The original manuscript highlights three practical rules: the evidence record precedes interpretation; at least some disfavored alternatives remain available to replay; and every durable self/world change is conditional on constitutional, validation and identity admission. For privacy, the word preserve refers to authorized bounded retention, not indefinite accumulation.

\subsection{Algorithm 2: idle review cycle, replay-dominant}
\begin{lstlisting}
inputs: shared C, E, I, W, f0, M, Q, phase-tagged log L
while no higher-priority external request and budget_available:
    # 1. Select evidence by failure, novelty, surprise and conflict
    batch = replay_sample(L, Q,
            priority={failure, novelty, counterfactual_gain,
                      high_error, contradiction})
    # 2. Check original labels and provenance, re-segment if needed
    seq = check_and_resegment(batch)
    if not cross_validate_phase_labels(seq):
        mark_unreliable(seq)
        continue
    # 3. Attribute causes of outcomes
    cause = attribute(seq, model=f0, with_uncertainty=True)
    # 4. Evaluate counterfactual alternatives
    alternatives = simulate(f0, seq, interventions=cause)
    # 5. Abstract evidence-grounded skills
    skills = abstract_and_calibrate(seq, alternatives)
    # 6. Build a specific versioned proposal
    proposal = build_evolution_proposal(
        kind="self-or-world", diff=skills,
        evidence_ids=seq.ids, risk=estimate_risk(skills))
    # 7. Validate four independent conditions
    gates = {safety_check(proposal),
             capability_regression_check(proposal),
             consistency_check(proposal),
             stakeholder_value_check(proposal)}
    if not all(gates):
        log_rejection(proposal)
        continue
    # 8. Project to feasible state and revalidate the final diff
    p2 = project(proposal, C)
    if not validate(p2) or not identity_ok(p2, E):
        request_arbitration_or_reject(p2)
        continue
    # 9. Only the slow evolution clock may commit durable changes
    commit_or_seek_approval(p2,
        graded_by={impact_radius, reversibility, permissions})
    audit_version(p2, source_evidence=seq.ids)
    # 10. Monitor goals and consolidate only under policy
    update_gap_metrics(deltaS, deltaW)
    log_phase_and_admission_metrics(phase="phi2")
    if safe_to_consolidate(deltaS, deltaW):
        consolidate_with_retention_and_privacy_constraints(M, Q)
\end{lstlisting}

\section{Relationship to the predecessor and the source version}\label{app:source-correspondence}
The original Chinese paper is a continuation of \citet{tong2026unified}. Its axioms A1--A6 and H1--H10 are inherited, while the phase variable, micro-epoch contract, duty-cycle model, T1--T5, and TP1--TP12 are introduced as new constructs. The previous paper's use of two channels is broadened from potential exclusive scheduling to time-division occupancy, with \(\omega\in\{0,1\}\) as an endpoint case. The present version revises the mathematical claims rather than copying their overstrong statements: deterministic exact-gradient descent is proved, stochastic descent is reported with variance and assumptions, and cross-channel consistency is treated as a testable semantic property plus a procedural admission invariant.

The content of the fourteen original chapters is represented throughout the expanded paper as follows: Introduction and neuroscience basis (Sections 1--2), temporal layer (Section 3), formal results (Section 4), detailed busy/idle procedures (Sections 5--6), integration argument (Section 7), engineering contracts and road map (Section 8), all original TP hypotheses and phase-conditioned evaluation (Section 9), comparison and theory (discussion), ethics and new temporal risks, twelve open problems, limitations, and conclusion. This translation is editorially reorganized to accommodate new experiments; it is not a word-for-word certified translation.

\section{Version and reference provenance}
The editorial source is the supplied ODT manuscript (v1.0). The English translation and additional E1--E5/Tier-2 experimental discussion were drafted from the source and the existing code/result files on 4 November 2026. Bibliographic records were checked against primary publishers, conference proceedings, PubMed, Crossref and OpenAlex where available. The accompanying audit report lists all originally supplied reference keys, their URLs and corrections. No manuscript content should be interpreted as independent experimental verification of claims made only in cited neuroscience literature.

\section{Proofs and formal qualifications}\label{app:proof}
\paragraph{Proof of Proposition~\ref{prop:desc}.}
Write \(y=\Proj(x-\eta\nabla\Phi(x))\), \(d=y-x\). Optimality of Euclidean projection onto a closed convex set gives
\[
 \langle x-\eta\nabla\Phi(x)-y,\;x-y\rangle\leq0,
 \quad\Longrightarrow\quad
 \langle\nabla\Phi(x),d\rangle\leq-\frac{\|d\|^2}{\eta}.
\]
The smoothness descent lemma gives
\[
 \Phi(y)\leq \Phi(x)+\langle\nabla\Phi(x),d\rangle
  +\frac{L}{2}\|d\|^2
 \leq \Phi(x)-\left(\frac1\eta-\frac L2\right)\|d\|^2.
\]
When the two directions agree with the exact gradient, the duty-cycle weighted direction is also exact. This proof does not invoke unbiasedness. It requires an exact gradient for the pointwise statement.

\paragraph{Proof of Proposition~\ref{prop:stoch}.}
Let \(g=\nabla\Phi(x)\), \(\xi=\omega(G_{\rm on}-g)+(1-\omega)(G_{\rm off}-g)\), and \(y=x-\eta(g+\xi)\). Smoothness yields
\[
 \Phi(y)\leq\Phi(x)-\eta\langle g,g+\xi\rangle
   +\frac{L\eta^2}{2}\|g+\xi\|^2.
\]
Take conditional expectations and use \(\E[\xi\mid x]=0\),
\(\E\|\xi\|^2\leq\omega^2v_{\rm on}+(1-\omega)^2v_{\rm off}\) by conditional independence. The resulting gradient coefficient is
\(-\eta(1-L\eta/2)\leq-\eta/2\) for \(\eta\leq1/L\). This bound applies to the \emph{ungated, unconstrained} update. The acceptance gate can bias the selection of candidates and is not covered.

\paragraph{Proof of Lemma~\ref{lem:gain}.}
Since \(\Proj(x)=x\) for \(x\in\mathcal C\), nonexpansiveness gives
\(\|\Proj(x-\eta g)-\Proj(x)\|\leq\eta\|g\|\).
The two step-size bounds follow from convex combination and substitution of
\(\gamma_{\rm out}=\rho\gamma_{\rm in}\).
Neither step alone proves global convergence or cross-channel semantic consistency.

\paragraph{A simple resource/content counterexample.}
Suppose broadcast utility requires urgency \(u>0.7\) while capacity is inversely related to load \(\ell>0.7\). A coupled scalar \(m=(u+1-\ell)/2\) may be below the broadcast threshold even for urgent items (for example \(u=0.8,\ell=0.9\) gives \(m=0.45\)). A controller with separate urgency thresholding broadcasts this item. This establishes an \emph{existence} counterexample for a particular coupling, not an if-and-only-if characterization of scheduling architectures.

\section{Complete hypothesis and evidence map}\label{app:hyp}
\begin{table}[h]
\centering\small
\caption{Twelve original TP hypotheses, now distinguished from the strength of available evidence. A saved smoke-test metric or toy counterexample is not a general validation.}
\begin{tabular}{p{0.09\linewidth} p{0.48\linewidth} p{0.32\linewidth}}
\toprule
ID & Proposed testable prediction & Evidence available here\\
\midrule
TP1 & Phase occupancy and winning item switch with request pressure & API prototype, assignment-driven quick test\\
TP2 & Pruning shadow alternatives harms later proposal quality & API prototype, engineered quality bonus\\
TP3 & Removing interleaved shadow work harms long-run error recovery & Not tested\\
TP4 & Changing key operation order degrades accuracy/repair & API prototype, programmed confidence effect\\
TP5 & Non-unit gain ratio can optimize time-to-threshold & E3 toy simulation, conditional on noise law\\
TP6 & Shared model/validation improves cross-phase skill transfer & Not tested\\
TP7 & Uncoordinated transitions impair stable admission & Not tested\\
TP8 & Coupling content and resource controls misallocates gain & E4 constructed counterexample\\
TP9 & Missing error logs lower replay-based failure elimination & E5 probabilistic simulation\\
TP10 & Premature consolidation loses useful experiences & Not tested\\
TP11 & Phase-stratified scoring changes apparent framework ranking & Not tested\\
TP12 & Removing operators yields predicted location-specific failures & Not tested\\
\bottomrule
\end{tabular}
\end{table}

\section{Visualization provenance and reporting precision}\label{app:figs}
Figures 1--3 show conceptual architecture and scheduling mechanisms. Figures 4--5 plot saved full-run E1--E4 simulation values, with Figure 5(a) showing the entire original twelve-point E3 gain grid rather than a stylized curve. Figure 6 uses the original E5 simulation code and the same 64 seeded random streams to reconstruct all mode-wise trajectories. The reproduced fitted slope and rare/common ratio match the stored E5 JSON to numerical precision. Figure 6 displays means and normal-approximation 95\% confidence intervals (mean \(\pm1.96\) standard errors) over the original seeds. Numeric values and the per-alpha curves are also exported in the figure-data supplementary CSV and compressed seed arrays. All six embedded figures remain PDF vectors with no raster image layers.

\section{Experimental specifications and reproduction}\label{app:repro}
\begin{table}[h]
\centering\small
\caption{Full-run settings read from experiment scripts and saved JSON. A seed denotes a synthetic repeat, not an independently sampled real participant or dataset.}
\begin{tabular}{p{0.1\linewidth} p{0.79\linewidth}}
\toprule
Exp. & Configuration\\
\midrule
E1 & 64 seeds; \(d=20\), 3,000 steps; \(\eta_{\rm on}=\eta_{\rm off}=0.5\), \(\sigma_{\rm on}=0.6\), \(\sigma_{\rm off}=0.15\), weights \(0,0.25,0.5,0.75,1\); 500 deterministic inequality checks.\\
E2 & 64 seeds; scalar tilt strengths \(0,0.25,\ldots,2\); weights \(1,0.75,0.5,0.25\); report two terminal objective families separately.\\
E3 & 64 seeds; \(d=20\); \(\bar\omega=0.7\); 12 log-spaced \(\rho\in[0.5,6]\); \(\kappa\in\{0,0.5,1\}\); target potential \(0.3\); horizon 8,000.\\
E4 & 64 seeds; 20,000 synthetic task draws per correlation setting; urgency/load correlation \(0,0.3,0.6,0.9\); content threshold \(0.5\); logistic response with prescribed gain requirement.\\
E5 & 64 seeds; replay budget \(T=30\times60=1,800\); eight failure modes with rates \(0.01\) to \(0.0008\); deletion \(\alpha=0,0.1,\ldots,0.9\); evidence thresholds \(1,3,5\).\\
Tier 2 & Saved quick run: 15 generated tasks, five rounds, 16 workers, frozen \texttt{api/qwen3-8b} backend, logically scheduled micro-epochs. Independent true task correctness was not scored.\\
\bottomrule
\end{tabular}
\end{table}

The numerical tables in the paper are transcribed from \texttt{experiments/results/e1\_prop1\_full.json} through \texttt{e5\_tp9\_logdrop\_full.json} and \texttt{tier2\_parallel\_quick.json}. To regenerate the Tier-1 plots, install Python 3.10+ with \texttt{numpy}, \texttt{scipy}, and \texttt{matplotlib} and run the commands below from the included experiment directories. CPU execution is sufficient.
\begin{quote}
\small\ttfamily
cd experiments/tier1\_simulation\\
python run\_all.py --quick\\
python run\_all.py
\end{quote}
For the API-backed prototype, serve a compatible frozen model, and from \texttt{experiments/tier2\_mvp} run:
\begin{quote}
\small\ttfamily
python run\_parallel.py --quick --backbone api\\
\phantom{python}\hspace*{0.4cm} --model qwen3-8b --workers 16
\end{quote}
The backend defaults to \texttt{http://localhost:8000/v1}. A separate \texttt{--base-url} option is available. Running with \texttt{--backbone mock} checks only the deterministic software pathway. Both backends use the same synthetic task generator.

\paragraph{Statistical interpretation.}
Where plotted, intervals are computed by the original scripts as mean \(\pm1.96\) standard errors across seeds (normal approximation) rather than bootstrap intervals. The E3 median threshold search and E5 fitted slope are point summaries; no confidence interval for the optimum or slope is claimed here. The original code reuses a seeded deterministic generator in certain comparisons, and not all random streams are paired across ablation arms. The supplement retains the source and JSON outputs so these choices can be audited.

\section{Design principles inherited from the Chinese manuscript}
\begin{table}[h]
\centering\small
\caption{The original T1--T5 design principles as engineering hypotheses. Neuroscientific observations motivate but do not entail them.}
\begin{tabular}{p{0.11\linewidth} p{0.31\linewidth} p{0.47\linewidth}}
\toprule
ID & Design choice & Testable consequence\\
\midrule
T1 & Interleave phase-tagged micro-epochs and retain shadow candidates & Retention/replay ablations alter later performance.\\
T2 & Allow asymmetric foreground/background gain & Gain ratio interacts with noise and learning speed.\\
T3 & Share backbone, verification and projection operators & Shared and independent variants differ in skill transfer/consistency.\\
T4 & Coordinate phase transitions; commit on a slower clock & Removing transition barriers changes invalid commits or recovery.\\
T5 & Separate content selection from resource-gain decisions & Coupling can misallocate compute on conflicting load/urgency inputs.\\
\bottomrule
\end{tabular}
\end{table}
The earlier manuscript also maps four self-views (constitutional, old, ideal, proposed new), four world-views, two data domains, three temporal perspectives, and eight cognitive functions into the same event stream. That mapping is a \emph{proposed data architecture}; the present experiments isolate the scheduling and admission pieces rather than evaluating all those constructs jointly.

\end{document}